\documentclass[journal, oneside]{IEEEtran}

\usepackage{mathtools}

\PassOptionsToPackage{bookmarks={false}}{hyperref}
\usepackage{graphicx}
\usepackage{lettrine}
\usepackage{booktabs}
\usepackage{amsmath}
\usepackage{enumerate}
\usepackage{bm}
\usepackage{graphicx}
\usepackage{epstopdf}
\usepackage{amsmath}
\usepackage{array}
\usepackage{amssymb}
\usepackage[square, comma, sort&compress, numbers]{natbib}
\usepackage{bm}
\usepackage{subfigure}
\usepackage{url}
\usepackage{epstopdf}

\usepackage{xcolor}
\usepackage{soul}

\newcolumntype{P}[1]{>{\raggedright\arraybackslash}p{#1}}

\DeclareMathOperator*{\argmax}{argmax}
\ifCLASSINFOpdf
\else
\fi
  \usepackage[caption=false,font=footnotesize]{subfig}

\usepackage{tikz}
\usepackage{textcomp}

\newcommand\copyrighttext{%
  \footnotesize \textcopyright 2019 IEEE. Personal use of this material is permitted.
  Permission from IEEE must be obtained for all other uses, in any current or future 
  media, including reprinting/republishing this material for advertising or promotional 
  purposes, creating new collective works, for resale or redistribution to servers or 
  lists, or reuse of any copyrighted component of this work in other works. 
  DOI: 10.1109/JSAC.2019.2951964}
\newcommand\copyrightnotice{%
\begin{tikzpicture}[remember picture,overlay]
\node[anchor=south,yshift=5pt] at (current page.south) {\fbox{\parbox{\dimexpr\textwidth-\fboxsep-\fboxrule\relax}{\copyrighttext}}};
\end{tikzpicture}%
}

\begin{document}

\title{Fault Location in Power Distribution Systems \\ via Deep Graph Convolutional Networks}

\author{
Kunjin Chen, Jun Hu, \emph{Member, IEEE}, Yu Zhang, \emph{Member, IEEE}, Zhanqing Yu, \emph{Member, IEEE}, and Jinliang He, \emph{Fellow, IEEE}

\thanks{
Manuscript received May 27, 2019; revised September 16, 2019; accepted October 5, 2019. This work was supported in part by National Key R\&D Program of China under Grant 2018YFB0904603, Natural Science Foundation of China under Grant 51720105004, State Grid Corporation of China under Grant 5202011600UJ, the Hellman Fellowship, and the Faculty Research Grant (FRG) of University of California, Santa Cruz. 

K.-J. Chen, J. Hu, Z.-Q. Yu, and J.-L. He are with the State Key Lab of Power Systems, Dept. of Electrical Engineering, Tsinghua University, Beijing 100084, P. R. of China. 

Y. Zhang is with the Dept. of Electrical and Computer Engineering, University of California, Santa Cruz, CA 95064, USA.

(Corresponding author email: hejl@tsinghua.edu.cn).
} 
}

\maketitle

\maketitle
\copyrightnotice
\vspace{-10pt}

\markboth{}%
{Shell \MakeLowercase{\textit{et al.}}: Bare Demo of IEEEtran.cls for Journals}


%Telephone: (800) 555--1212, Fax: (888) 555--1212}


\pagestyle{empty}  % no page number for the second and the later pages
\thispagestyle{empty} % no page number for the first page

\begin{abstract}
This paper develops a novel graph convolutional network (GCN) framework for fault location in power distribution networks. The proposed approach integrates multiple measurements at different buses while taking system topology into account. The effectiveness of the GCN model is corroborated by the IEEE 123 bus benchmark system. Simulation results show that the GCN model significantly outperforms other widely-used machine learning schemes with very high fault location accuracy. In addition, the proposed approach is robust to measurement noise and data loss errors. Data visualization results of two competing neural networks are presented to explore the mechanism of GCN’s superior performance. A data augmentation procedure is proposed to increase the robustness of the model under various levels of noise and data loss errors. Further experiments show that the model can adapt to topology changes of distribution networks and perform well with a limited number of measured buses.
\end{abstract}

\smallskip


% \fi
\IEEEpeerreviewmaketitle

\section{Introduction}

Distribution systems are constantly under the threat of short-circuit faults that would cause power outages. In order to enhance the operation quality and reliability of distribution systems, system operators have to deal with outages in a timely manner. Thus, it is of paramount importance to accurately locate and quickly clear faults immediately after the occurrence, so that quick restoration can be achieved. 

Existing fault location techniques in the literature can be divided into several categories, namely, impedance-based methods \cite{liao2011generalized, krishnathevar2012generalized, das2012distribution}, voltage sag-based methods \cite{pereira2009improved, lotfifard2011voltage, trindade2014fault}, automated outage mapping \cite{trindade2014fault, teng2014automatic, jiang2016outage}, traveling wave-based methods \cite{thomas2003fault, shi2018travelling}, and machine learning-based methods \cite{thukaram2005artificial, mora2007fault, aslan2017artificial, hosseini2018ami}. Impedance-based fault location methods use voltage and current measurements to estimate fault impedance and fault location. Specifically, a generalized fault location method for overhead distribution system is proposed in \cite{liao2011generalized}. Substation voltage and current quantities are expressed as functions of the fault location and fault resistance, thus the fault location can be determined by solving a set of nonlinear equations. To solve the multiple estimation problem, it is proposed to use estimated fault currents in all phases including the healthy phase to find the faulty feeder and the location of the fault \cite{krishnathevar2012generalized}. It is pointed out in \cite{majidi2018new} that the accuracy of impedance-based methods can be affected by factors including fault type, unbalanced loads, heterogeneity of overhead lines, measurement errors, etc.

When a fault occurs in a distribution system, voltage drops can occur at all buses. The voltage drop characteristics for the whole system vary with different fault locations. Thus, the voltage measurements on certain buses can be used to identify the fault location. For instance, calculated fault currents can be applied to each bus in the system, and the values of voltage drop on a small number of buses can be obtained by calculating the power flows. The fault location can then be determined by comparing measured and calculated values of voltage drop \cite{pereira2009improved, lotfifard2011voltage}. In \cite{trindade2014fault}, multiple estimations of fault current at a given bus are calculated using voltage drop measurements on a small number of buses, and a bus is identified as the faulty bus if the variance of the multiple fault current estimates takes the smallest value.

Automatic outage mapping refers to locating a fault or reducing the search space of a fault using information provided by devices that can directly or indirectly indicate the fault location. For example, when a fault occurs, if an automatic recloser is disconnected, smart meters downstream of the device would experience an outage. Smart meters downstream of the fault itself will also feature a loss of power. Thus, the search space of the fault can be greatly reduced if the geographic location of each smart meter is considered \cite{trindade2014fault}. Authors in \cite{teng2014automatic} proposed to use fault indicators to identify the fault location. Each fault indicator can tell whether the fault current flows through itself (it may also have the ability to tell the direction of the fault current). The location of the fault can then be narrowed down to a section between any two fault indicators. An integer programming-based method is proposed in \cite{jiang2016outage} to locate a fault using information from circuit breakers, automatic reclosers, fuses, and smart meters. Multiple fault scenarios, malfunctioning of protective devices, and missing notifications from smart meters are also taken into consideration. 

Traveling wave-based methods use observation of original and reflected waves generated by a fault. Specifically, different types of traveling wave methods include single-ended, double-ended, injection-based, reclosing transient-based, etc. The principle and implementation of single-ended and double-ended fault location with traveling waves are discussed in \cite{thomas2003fault}. The traveling wave generated by circuit breaker reclosing is used to locate faults in \cite{shi2018travelling}. In general, however, traveling wave-based methods require high sampling rates and communication overhead of measurement devices \cite{lotfifard2011voltage}. Systems such as the global positioning system (GPS) are required for time synchronization across multi-terminal signals.

Machine learning models are leveraged for fault location in distribution systems \cite{chen2016fault}. Using the spectral characteristics of post-fault measurements, data with feature extraction are fed into an artificial neural network (ANN) for fault location \cite{aslan2017artificial}. A learning algorithm for multivariable data analysis (LAMDA) is used in \cite{mora2007fault} to obtain fault location. Descriptors are extracted from voltage and current waveforms measured at the substation. Various LAMDA nets are trained for different types of faults. In \cite{thukaram2005artificial}, the authors first use support vector machines (SVM) to classify the fault type, and then use ANN to identify the fault location. Smart meter data serves as the input of a multi-label SVM to identify the faulty lines in a distribution system \cite{hosseini2018ami}.

The deployment of distribution system measurement devices or systems such as advanced metering infrastructure \cite{hosseini2018ami}, micro phasor measurement units \cite{farajollahi2018locating}, and wireless sensor networks \cite{hossan2018data} improves data-driven situational awareness for distribution systems \cite{zhou2018partial, chen2018learning}. There are two major challenges for fault location in distribution systems with the increased number of measurements available: first, traditional fault location methods are unable to incorporate the measurements from different buses in a flexible manner, especially when the losses of data are taken into consideration. Second, for traditional machine learning approaches, the topology of the distribution network is hard to model, let alone the possibility of topology changes.

Recent advances in the field of machine learning, especially deep learning, have gained extensive attentions from both academia and industry. One of the major developments is the successful implementation of convolutional neural networks (CNN) in a variety of image recognition-related tasks \cite{lecun2015deep}. While the measurements on different buses in a power distribution system are spatially distributed, it is hard to directly implement a CNN model that use such measurements as input. Nevertheless, when multiple buses in a distribution system become measurable, it is possible to treat the measurements as signals on a graph to which variants of traditional data analysis tools may be applicable \cite{shuman2013emerging, sandryhaila2013discrete}. As an extension of CNNs for data on graphs, graph convolutional networks (GCN) have been designed and implemented, such that the advantages of CNNs can be exploited for data residing on graphs \cite{bruna2014spectral, defferrard2016convolutional, gama2018convolutional}.

In this paper, a GCN model is proposed for fault location in distribution systems. Unlike existing machine learning models used for fault location tasks, the architecture of the proposed model preserves the spatial correlations of the buses and learns to integrate information from multiple measurement units. Features are extracted and composited in a layer-by-layer manner to facilitate the faulty bus classification task. We also design a data augmentation procedure to ensure that the model is robust to varied levels of noise and errors. In addition, the proposed model can be readily adapted or extended to various tasks concerning data processing for multiple measurements in modern smart grids. 

The organization of the rest of the paper is as follows: in Section II, the fault location task is formulated and the proposed GCN model is described in detail. We also introduce the IEEE 123 bus test case used in this paper. The effectiveness of the proposed GCN model is validated in Section III with extensive comparisons and visualizations. A data augmentation procedure for training robust models is introduced. The performance of the model under topology changes and on high impedance faults is evaluated. We also implement the GCN model on another distribution network test case and discuss several practical concerns. Finally, Section IV concludes the paper and points out some future works. 

\section{Fault Location Based on Graph Convolutional Networks}

In this section, we first give a brief description of the fault location task. Next, we will revisit idea of spectral convolution on graphs, and show how a GCN can be constructed based on that idea. Finally, we will present the test case of the IEEE 123 bus distribution system.

\subsection{Formulation of the Fault Location Task}

In this paper, we assume that the voltage and current phasor measurements are available at phases that are connected to loads. That is, for a given measured bus in a distribution system, we have access to its three-phase voltage and current phasors $(V_1, \theta^V_1, V_2, \theta^V_2, V_3, \theta^V_3, I_1, \theta^I_1, I_2, \theta^I_2, I_3, \theta^I_3) \in \mathbb{R}^{12}$. Values corresponding to unmeasured phases are set to zero. A data sample of measurements from the distribution system can then be represented as $\mathbf{X} \in \mathbb{R}^{n_o \times 12}$, where $n_o$ is the number of observed buses. We formulate the fault location task as a classification problem. More specifically, given a data sample matrix $\mathbf{X}_i$, the faulted bus $\tilde{y_i}$ is obtained by $\tilde{y}_i = f(\mathbf{X}_i)$, where $f$ is a specific faulty bus classification model. A fault is correctly located if $\tilde{y}_i = y_i$, where $y_i$ indicates the true faulty bus corresponding to $\mathbf{X}_i$. 

As the convolution operation of CNN is carried out in local regions within the input data, local features can be extracted, and complex structures within the data can be represented with the increase of convolution layers (for a detailed description of CNN, the readers may refer to \cite{Lawrence1997Face}). However, traditional CNN models can not be applied to signals on a distribution network as the inputs for CNN are supposed to be in Euclidean domains, such as images represented by values on regular two-dimensional grids and sequential data that is one-dimensional \cite{chen2016detection}. Thus, we introduce how a convolutional network can be constructed with signals on graphs hereinafter.

\subsection{Spectral Convolution on Graphs}

To be self-contained, we first present a brief introduction to spectral graph theory \cite{levie2017cayleynets}. Suppose we have an undirected weighted graph $\mathcal{G}=(\mathcal{V}, \mathcal{E}, \mathbf{W})$, where $\mathcal{V}$ is the set of vertices with $|\mathcal{V}|=n$, $\mathcal{E}$ is the set of edges, and $\mathbf{W} \in \mathbb{R}^{n \times n}$ is the weighted adjacency matrix. The unnormalized graph Laplacian of $\mathcal{G}$ is defined as $\mathbf{\Delta}_u = \mathbf{D} - \mathbf{W}$, where $\mathbf{D}$ is the degree matrix of the graph with diagonal entries $\mathbf{D}_{ii}=\sum_j\mathbf{W}_{ij}$. Then, the normalized graph Laplacian is given as 
\begin{equation}
\mathbf{\Delta} = \mathbf{D}^{-1/2}\mathbf{\Delta}_u\mathbf{D}^{-1/2} 
                  = \mathbf{I} - \mathbf{D}^{-1/2}\mathbf{W}\mathbf{D}^{-1/2},
\end{equation}
where $\mathbf{I}$ is the identity matrix. The eigendecomposition of the positive semi-definite symmetric matrix $\mathbf{\Delta}$ yields $\mathbf{\Delta} = \mathbf{\Phi}\mathbf{\Lambda}\mathbf{\Phi}^\top$, where $\mathbf{\Phi}=(\bm{\phi}_1, \dots, \bm{\phi}_n)$ are orthonormal eigenvectors of $\mathbf{\Delta}$, and $\mathbf{\Lambda} = \mathrm{diag}(\lambda_1, \dots, \lambda_n)$ is the diagonal matrix with corresponding ordered non-negative eigenvalues $0 = \lambda_1 \leq \lambda_2 \leq \dots \leq \lambda_n$. Note that the smallest eigenvalue $\lambda_1$ equals zero with the eigenvector $\bm{\phi}_1=(\frac{1}{\sqrt{n}}, \cdots, \frac{1}{\sqrt{n}})$. 
By analogy with the Fourier transform in Euclidean spaces, graph Fourier transform (GFT) can be defined for weighted graphs using the orthonormal eigenvectors of $\mathbf{\Delta}$ \cite{hammond2011wavelets}. For a signal $\mathbf{f} \in \mathbb{R}^n$ on the vertices of graph $\mathcal{G}$ (each vertice has one value in this case), GFT is performed as $\mathbf{\hat{f}} = \mathbf{\Phi}^\top\mathbf{f}$ while the inverse GFT is $\mathbf{f} = \mathbf{\Phi}\mathbf{\hat{f}}$. Further, we can conduct convolution on graphs in the spectral domain also by analogy with convolution on discrete Euclidean spaces facilitated by Fourier transform. That is, spectral convolution of two signals $\mathbf{g}$ and $\mathbf{f}$ is defined as
\begin{equation}
\begin{aligned}
\mathbf{g} * \mathbf{f} = \mathbf{\Phi}\left((\mathbf{\Phi}^\top \mathbf{g}) \circ (\mathbf{\Phi}^\top \mathbf{f})\right) 
                        = \mathbf{\Phi} \, \mathrm{diag}(\hat{g}_1, \dots, \hat{g}_n) \mathbf{\Phi}^\top\mathbf{f},
\end{aligned}
\end{equation}
where $\circ$ indicates element-wise product between two vectors. Filtering of signal $\mathbf{f}$ by spectral filter $\mathbf{\mathbf{B}} = \mathrm{diag}(\bm{\beta})$ with $\bm{\beta} \in \mathbb{R}^n$ can then be expressed as $\mathbf{\Phi} \mathbf{\mathbf{B}} \mathbf{\Phi}^\top \mathbf{f}$. One major drawback of this formulation, however, is that the filters are not guaranteed to be spatially localized, which is a crucial feature of CNNs for data in Euclidean spaces, since localized filters are able to extract features from small areas of interest instead of the whole input. Using filters $h_{\bm{\alpha}}(\mathbf{\Lambda})$ that are smooth in spectral domain can bypass such an issue \cite{henaff2015deep, levie2017cayleynets}. For example, consider using a polynomial approximation
\begin{equation}
h_{\bm{\alpha}}(\mathbf{\Lambda}) = \sum^{K}_{k=0}{\alpha_{k}\mathbf{\Lambda}^k},
\end{equation}
where $\bm{\alpha} = (\alpha_1, \dots, \alpha_K)$ is the vector of coefficients to be learned for the filters and $K$ is the degree of the polynomials. Further, in order to stabilize the training of the polynomial filters, the truncated Chebyshev polynomial expansion of $h_{\bm{\alpha}}(\mathbf{\Lambda})$ is introduced \cite{hammond2011wavelets, defferrard2016convolutional}. Specifically, expansion of $h_{\bm{\alpha}}(\mathbf{\Lambda})$ using Chebyshev polynomials $\mathit{T}_k(\tilde{\mathbf{\Lambda}})$ up to order $K$ can be expressed as 
\begin{equation}
h_{\bm{\alpha}}(\mathbf{\Lambda}) = \sum^{K}_{k=0}{\alpha_k \mathit{T}_k(\tilde{\mathbf{\Lambda}})},
\end{equation}
where $\tilde{\mathbf{\Lambda}} = 2\mathbf{\Lambda}/\lambda_n - \mathbf{I}$. The recursive formulation of the filtering process based on Chebshev polynomials is introduced in \cite{defferrard2016convolutional}, which takes the form $\mathit{T}_k(x) = 2x\mathit{T}_{k-1}(x) - \mathit{T}_{k-2}(x)$ with $\mathit{T}_0 = 1$ and $\mathit{T}_1 = x$. Since $\mathbf{\Delta}^k = (\mathbf{\Phi} \mathbf{\Lambda} \mathbf{\Phi}^\top)^k = \mathbf{\Phi} \mathbf{\Lambda}^k \mathbf{\Phi}^\top$, the filtering process $\mathbf{\Phi} h_{\bm{\alpha}}(\mathbf{\Lambda}) \mathbf{\Phi}^\top \mathbf{f}$ can be expressed as
\begin{equation}
\mathbf{\Phi} h_{\bm{\alpha}}(\mathbf{\Lambda}) \mathbf{\Phi}^\top \mathbf{f} = h_{\bm{\alpha}}(\mathbf{\Delta}) \mathbf{f} =
\sum^{K}_{k=0}{\alpha_k \mathit{T}_k(\tilde{\mathbf{\Delta}})} \mathbf{f},
\end{equation}
where $\tilde{\mathbf{\Delta}} = 2\mathbf{\Delta}/\lambda_n - \mathbf{I}$. Consequently, with $\mathbf{d}_0 = \mathbf{f}$ and $\mathbf{d}_1 = \tilde{\mathbf{\Delta}}\mathbf{f}$, we can recursively calculate $\mathbf{d}_k = 2\tilde{\mathbf{\Delta}}\mathbf{d}_{k-1} - \mathbf{d}_{k-2}$, and the filtering operation $h_{\bm{\alpha}}(\mathbf{\Delta}) \mathbf{f} = [\mathbf{d}_{0}, \cdots, \mathbf{d}_{K}]\bm{\alpha}$ has a computational complexity of $\mathcal{O}(K|\mathcal{E}|)$ considering the sparsity of $\mathbf{\Delta}$ \cite{defferrard2016convolutional}. In addition, because the Chebyshev polynomials are truncated to the $K$th order, the filter is $K$-hop localized with respect to the connections embodied in $\mathbf{\Delta}$. To this end, GCN can be implemented with the aforementioned spectral convolution on graphs.

\subsection{GCN Approach for Fault Location}

\begin{figure*}[!t]
\setlength{\abovecaptionskip}{0pt}
\centering
\includegraphics[width=11.5cm]{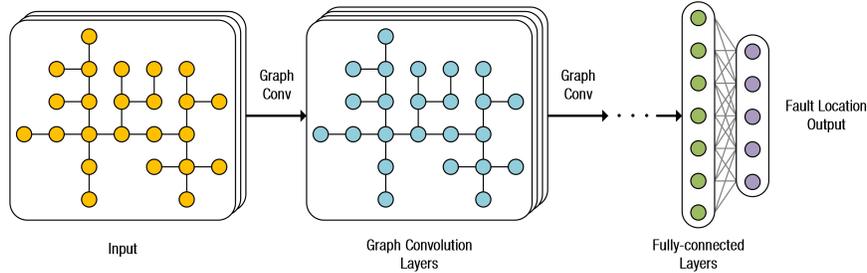}
\caption{The structure of the GCN model. Several graph convolution layers are followed by two fully-connected layers.}
\label{model}
\end{figure*}

The GCN model applied to the fault location task is illustrated in Fig. \ref{model}. The input $\mathbf{X}$ is passed through $L_c$ graph convolution layers and $L_f$ fully-connected layers followed by a softmax activation function. Specifically, the $j$th feature map of a graph convolution layer is calculated as
\begin{equation}
\mathbf{y}_j = \sum_{i=1}^{N_{in}}{h_{\bm{\alpha}_{i,j}}(\mathbf{\Delta})\mathbf{x}_i},
\end{equation}
where $\mathbf{x}_i \in \mathbb{R}^n$ is the $i$th input feature map, $\bm{\alpha}_{i,j} \in \mathbb{R}^K$ is the trainable coefficients, and $N_{in}$ is the number of filters of the previous layer. With $N_{out}$ filters in the current layer, a total of $N_{in}N_{out}K$ parameters are trainable in this layer. In particular, $N_{in}=12$ for the first layer of the model. The output of the last graph convolution layer is flattened into a vector and passed to the fully-connected layers. The index of the predicted faulty bus, $\tilde{y}$, can be obtained as $\tilde{y} = \argmax_i a_i$, where $a_i$ is the $i$th activation of the last fully-connected layer.

The weighted adjacency matrix is constructed based on the physical distance between the nodes. First, the distance matrix $\mathbf{S}\in\mathbb{R}^{n\times n}$ is formed with $\mathbf{S}_{ij}$ being the length of the shortest path between bus $i$ and bus $j$. We then sort and keep the smallest $K_n$ values in each row of $\mathbf{S}$ to obtain $\tilde{\mathbf{S}}\in \mathbb{R}^{n\times K_n}$ and calculate $\sigma_S = \sum_i{\tilde{\mathbf{S}}_{iK_n}}/n$ (we have $\tilde{\mathbf{S}}_{ij} \leq \tilde{\mathbf{S}}_{ik}$ for $j<k$). Matrix $\tilde{\mathbf{W}}\in \mathbb{R}^{n\times K_n}$ is then constructed with $\tilde{\mathbf{W}}_{ij} = e^{-\tilde{\mathbf{S}}_{ij}^2/\sigma_S^2}$. By restoring the positional correspondence of $\tilde{\mathbf{W}}_{ij}$ to bus $i$ and bus $j$, the weighted adjacency matrix $\mathbf{W}\in \mathbb{R}^{n\times n}$ can be obtained. We can thus proceed to compute $\mathbf{D}$ and finally obtain $\mathbf{\Delta}$ according to (1).

\subsection{The IEEE 123 Bus Distribution System Test Case}

\begin{figure*}[!t]
\setlength{\abovecaptionskip}{0pt}
\centering
\includegraphics[width=11.5cm]{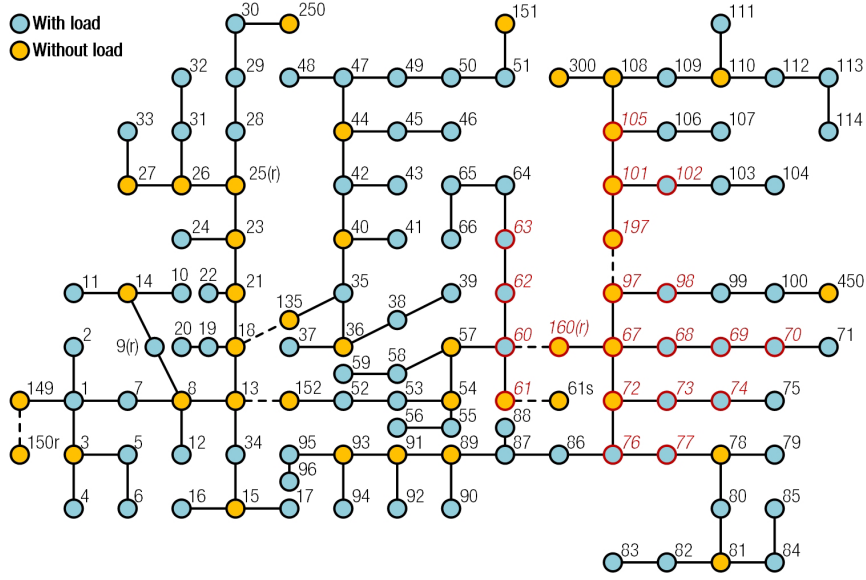}
\caption{An illustration of the IEEE 123 bus system. It is assumed that voltage and current phasors of PQ buses (connected to loads) are measured. As an example, the 20 buses that are closest (in distance) to bus 67 are highlighted with red color and italic numbers. Normally closed switches are represented by dashed lines.}
\label{IEEE123}
\end{figure*}

The IEEE 123 bus test case is used to carry out the task of fault location in distribution systems in this paper \cite{kersting2001radial}. The overall topology of the distribution system is illustrated in Fig. \ref{IEEE123}. Note that the topology is only used to indicate the connections of the buses rather than their geometrical locations. Specifically, there are 128 buses in the system (cf. Fig. \ref{IEEE123}), 85 of which are connected to loads. Most of those loads are only connected to a single phase. Bus pairs (149, 150r), (18, 135), (13, 152), (60, 160(r)), (61, 61s), and (97, 197) are connected by normally closed switches. In addition, regulators are installed at buses 9, 25, and 160. 

\begin{figure}[!t]
\setlength{\abovecaptionskip}{0pt}
\centering
\includegraphics[width=6.3cm]{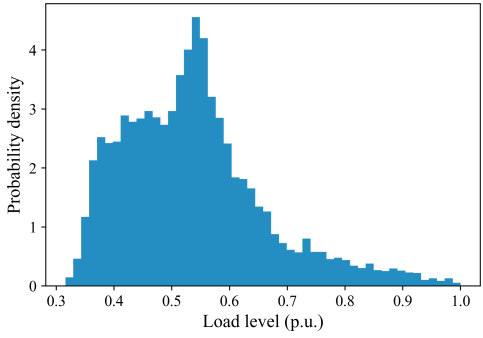}
\caption{The probability density of load level for the distribution system.}
\label{density}
\end{figure}

In order to generate the training and test datasets, faults are simulated for all buses in the system. Three types of faults are considered, namely, single phase to ground, two phase to ground, and two phase short-circuit. The faults have the resistance ranging from 0.05 $\Omega$ to 20 $\Omega$. The load level of the system varies between 0.316 and 1. In order Fig. \ref{density} shows the discrete probability density function (PDF) with 50 equal-length load level intervals. The PDF is obtained from the annual load curve of the system. We randomly sample one value from the load level distribution and set all loads in the system to the same level. The simulations are implemented by the OpenDSS software \cite{dugan2011open}. The voltage and current phasors are measured during the fault. We obtain the training and test datasets used for training and evaluating the fault location models. 

We generate 20 data samples for each fault type at each bus. As a result, a total of 13520 data samples are generated for both the training and test datasets. We consider buses connected with normally closed switches or regulators as a single bus. Thus, there are a total of 119 faulty buses to be classified; i.e., 119 class labels for the classification task.

For the implementation of the GCN model\footnote{The implementation of GCN in this paper is based on the implementation in \cite{defferrard2016convolutional}; see https://github.com/mdeff/cnn\_graph}, instead of using $n_o \times 12$ as the size of the input of the model, we expand $\mathbf{X}$ to include all 128 buses, i.e., each input data sample has a size of $128 \times 12$. As a result, each sample matrix has 1536 entries, 380 of which have measured values. For the non-measured buses, we set the corresponding values to be zero. The same measured quantity is run through the standardization process; i.e., subtracting the mean and dividing by the standard deviation.

\section{Results and Discussion}

In this section, we report the performance of GCN for fault location tested in the IEEE 123 bus benchmark system. Comparisons with baseline models are provided in detail. We also visualize the hidden features of samples in the test dataset to demonstrate that the proposed GCN model is able to learn more robust representations from data. 

\subsection{Implementation Details and Baseline Models}

The hyper-parameters of the GCN model implemented in this paper are determined using 10\% of the training dataset as the validation dataset. Specifically, the model has 3 graph convolution layers (all with 256 filters) followed by 2 fully-connected layers (with 512 and 256 hidden nodes). $K_n$ is set to 20 and the values of $K$ for the graph convolution layers are 3, 4, and 5, respectively. The two fully-connected layers have a dropout rate of 0.5. The Adam optimizer with an initial learning rate of 0.0002 is used to train the model for 400 epochs (i.e., each data sample is used 400 times for training) and a mini-batch size of 32. We use Tensorflow in Python to implement the GCN model. When trained with a Titan Xp GPU, the GCN model takes less than 2 hours to train, and the time used to test each sample in the test dataset is less than 0.5 ms.

\begin{figure}[!t]
\centering
\subfigure[]{
\includegraphics[width=3.2cm]{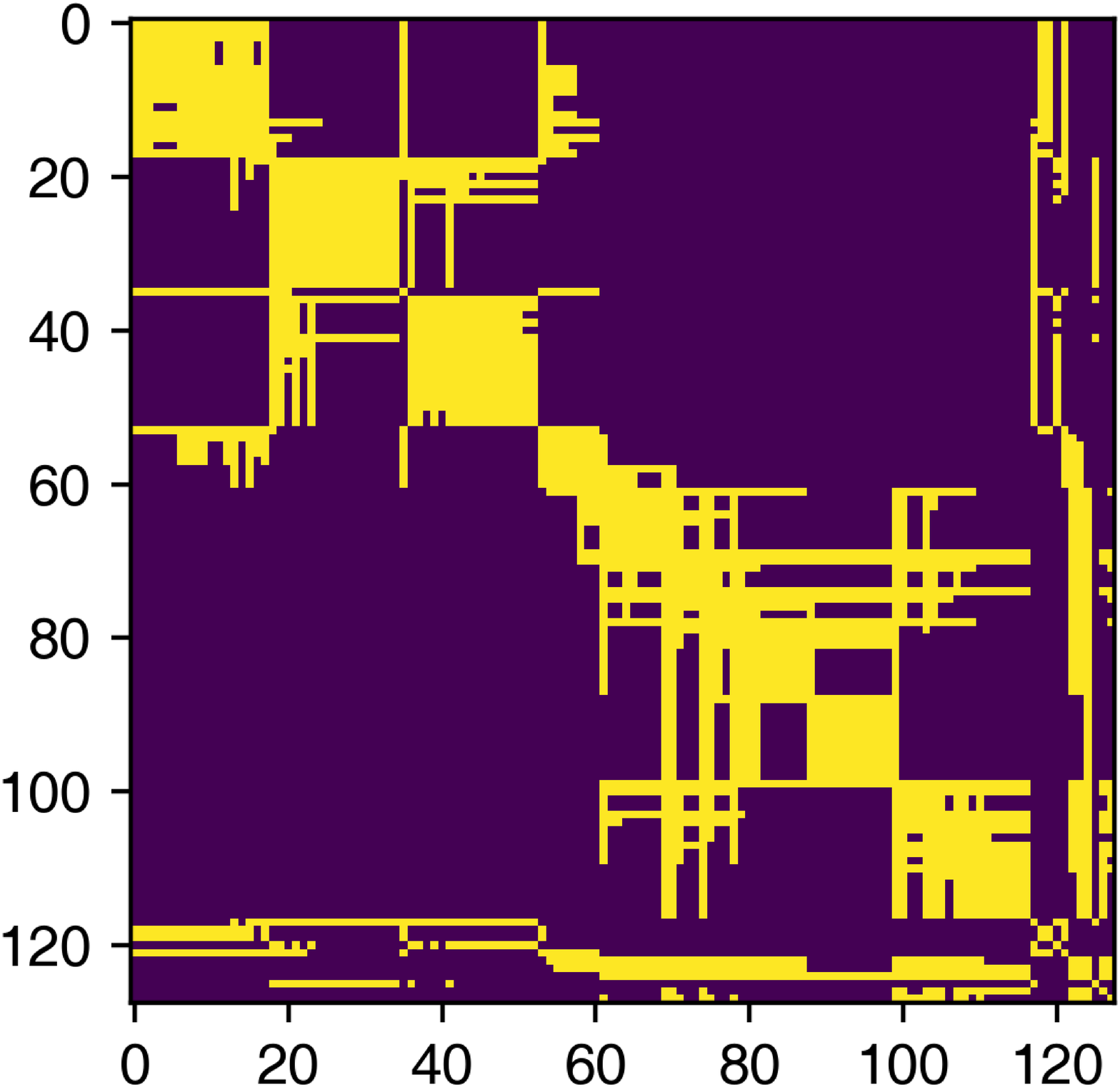}
%\caption{fig1}
}
\quad
\subfigure[]{
\includegraphics[width=3.2cm]{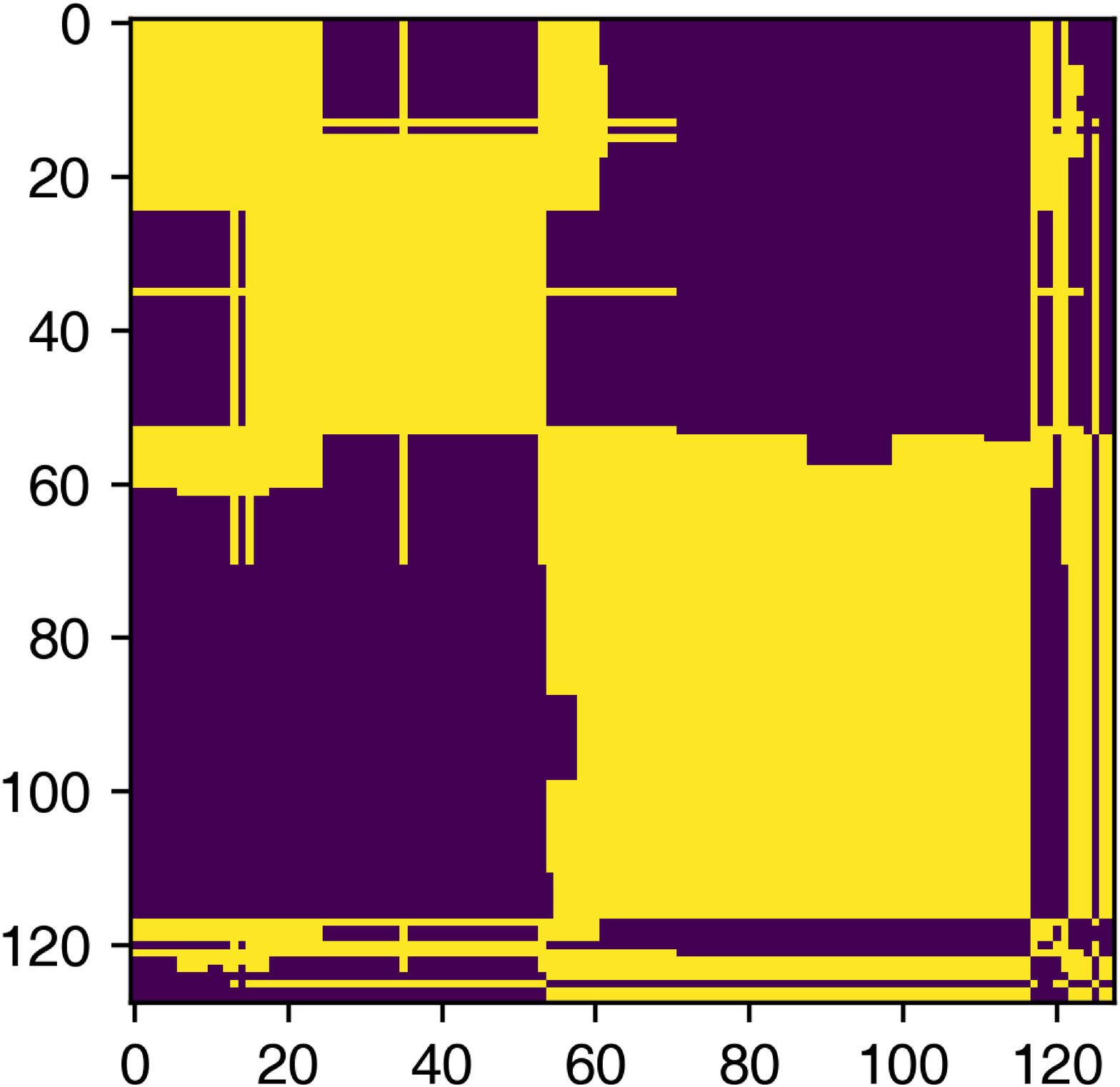}
}
\quad
\subfigure[]{
\includegraphics[width=3.2cm]{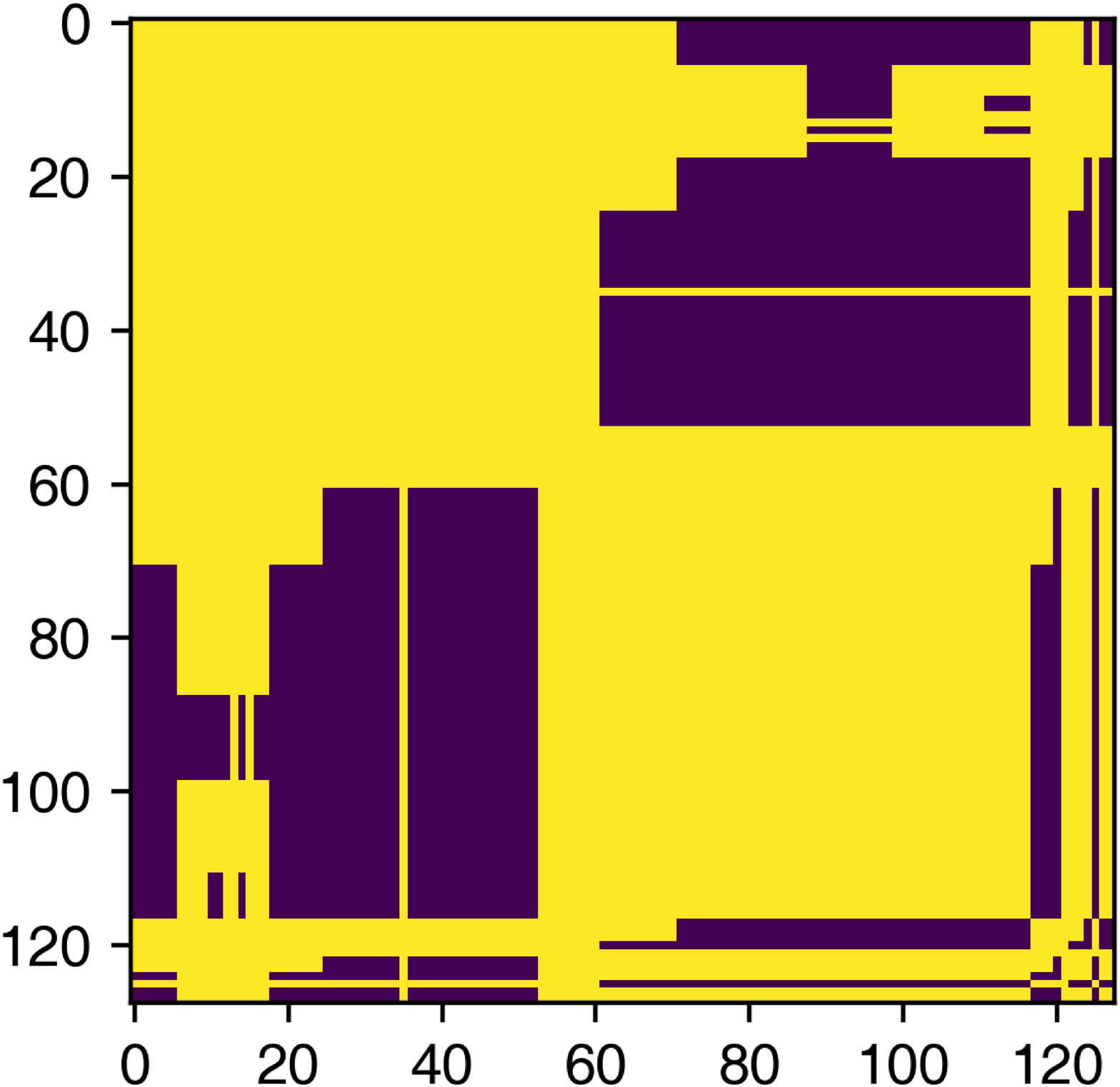}
}
\quad
\subfigure[]{
\includegraphics[width=3.2cm]{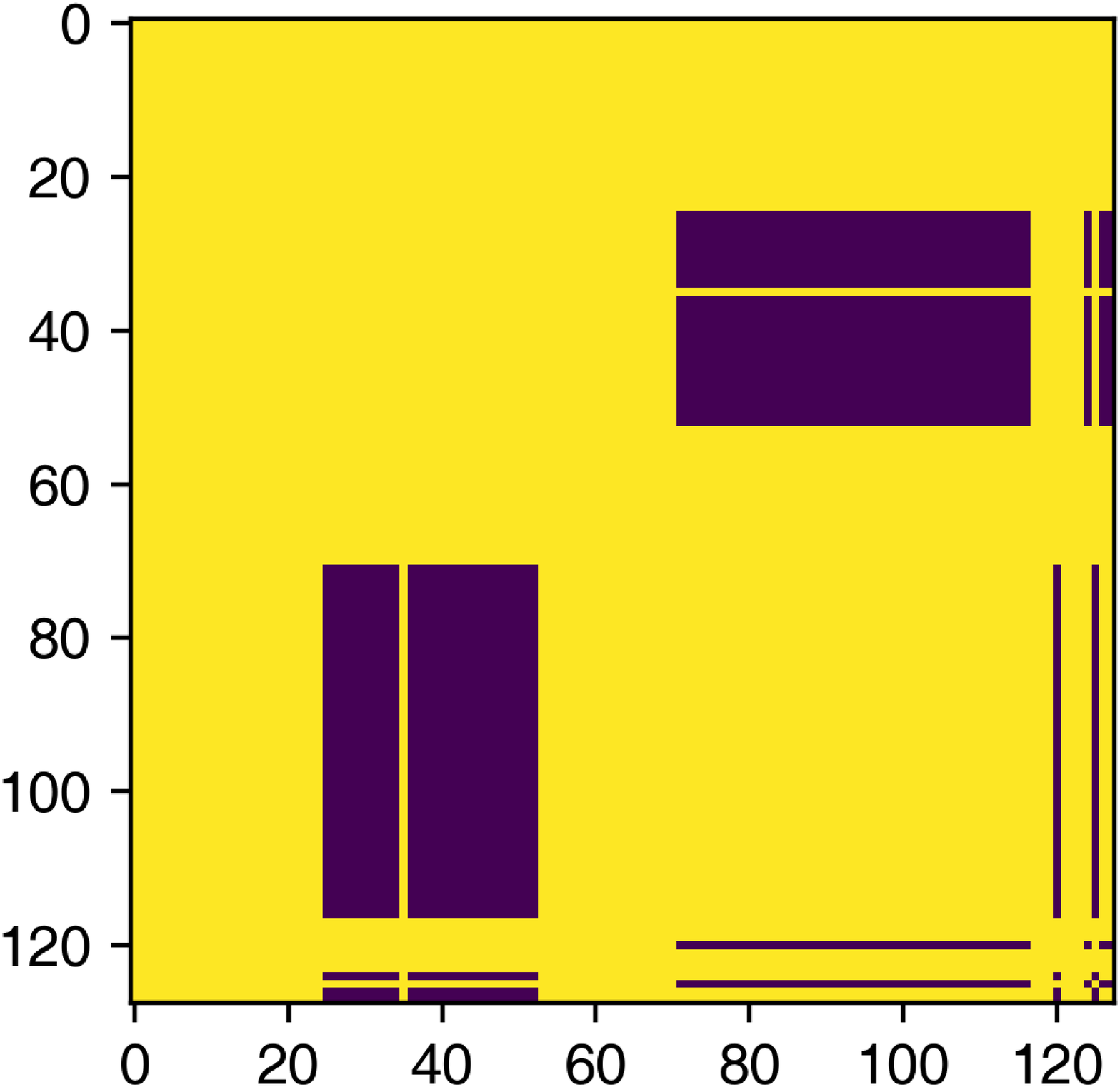}
}
\caption{Visualization of non-zero entries (yellow) in $\mathbf{\Delta}^m$ when $K_n=20$ (20 nearest buses of bus 67 are red colored as shown in Fig. \ref{IEEE123}): (a) $m=1$, (b) $m=2$, (c) $m=3$, and (d) $m=4$.}
\label{vis1}
\end{figure}

\begin{figure}[!t]
\centering
\subfigure[]{
\includegraphics[width=3.7cm]{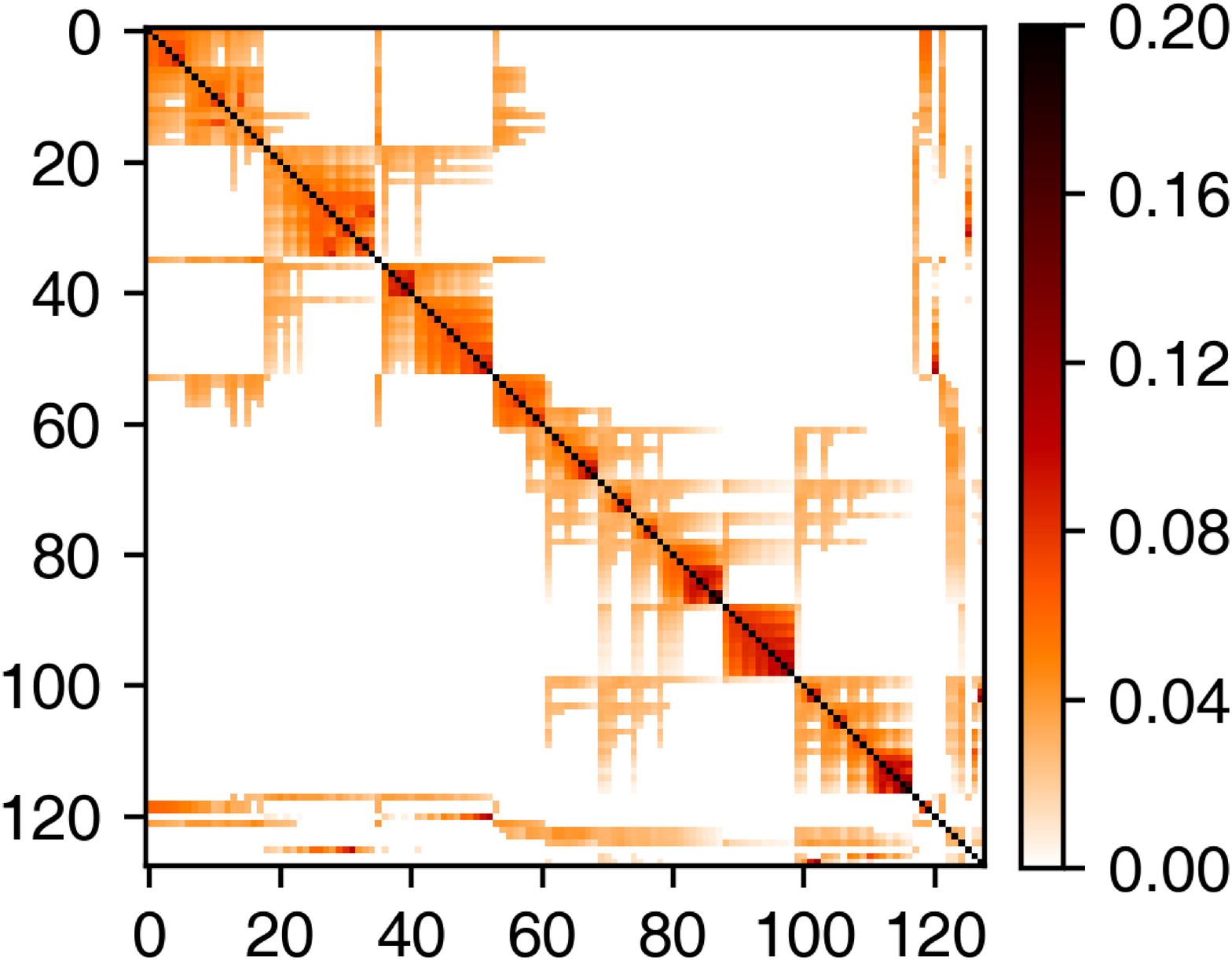}
%\caption{fig1}
}
\quad
\subfigure[]{
\includegraphics[width=3.7cm]{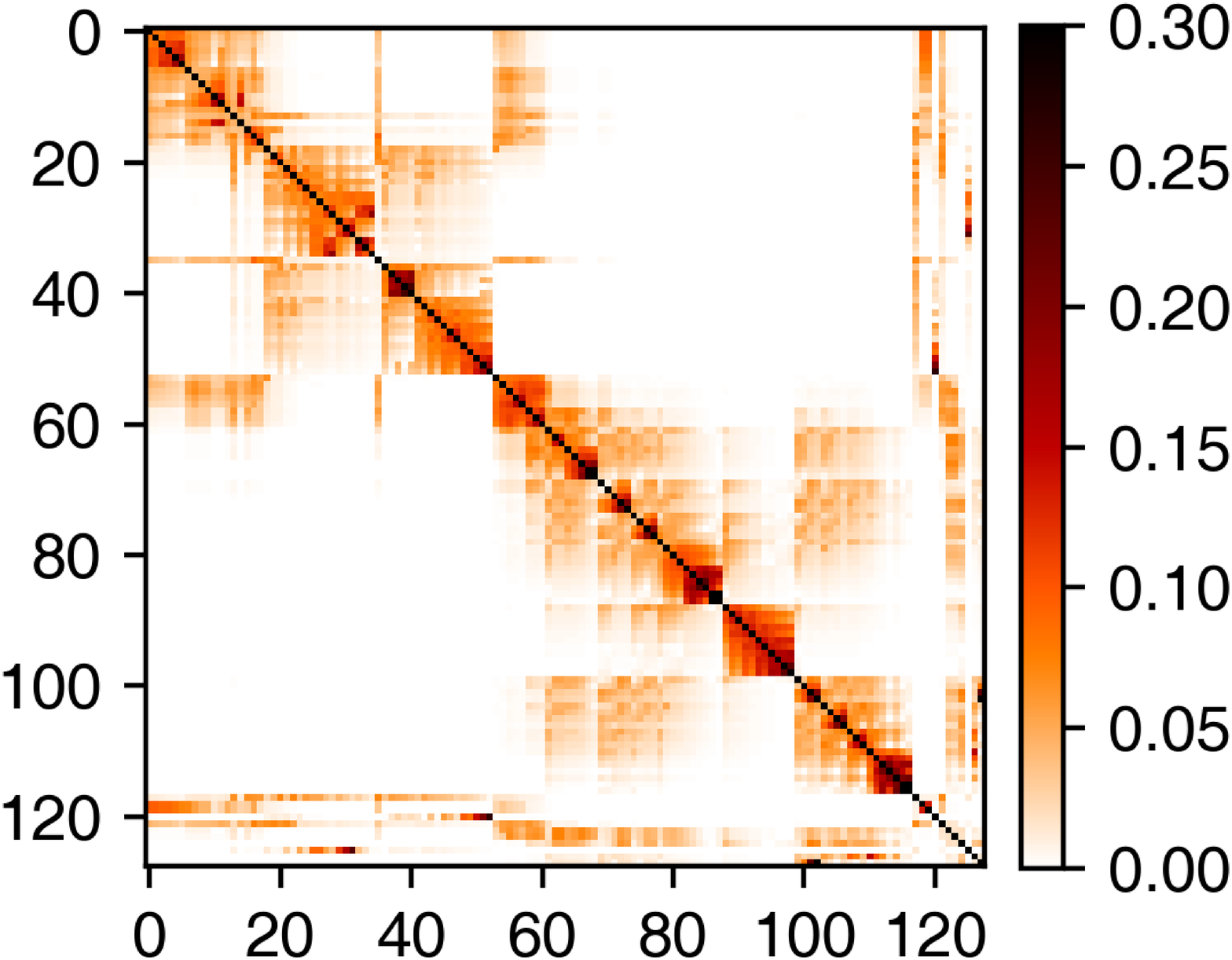}
}
\quad
\subfigure[]{
\includegraphics[width=3.7cm]{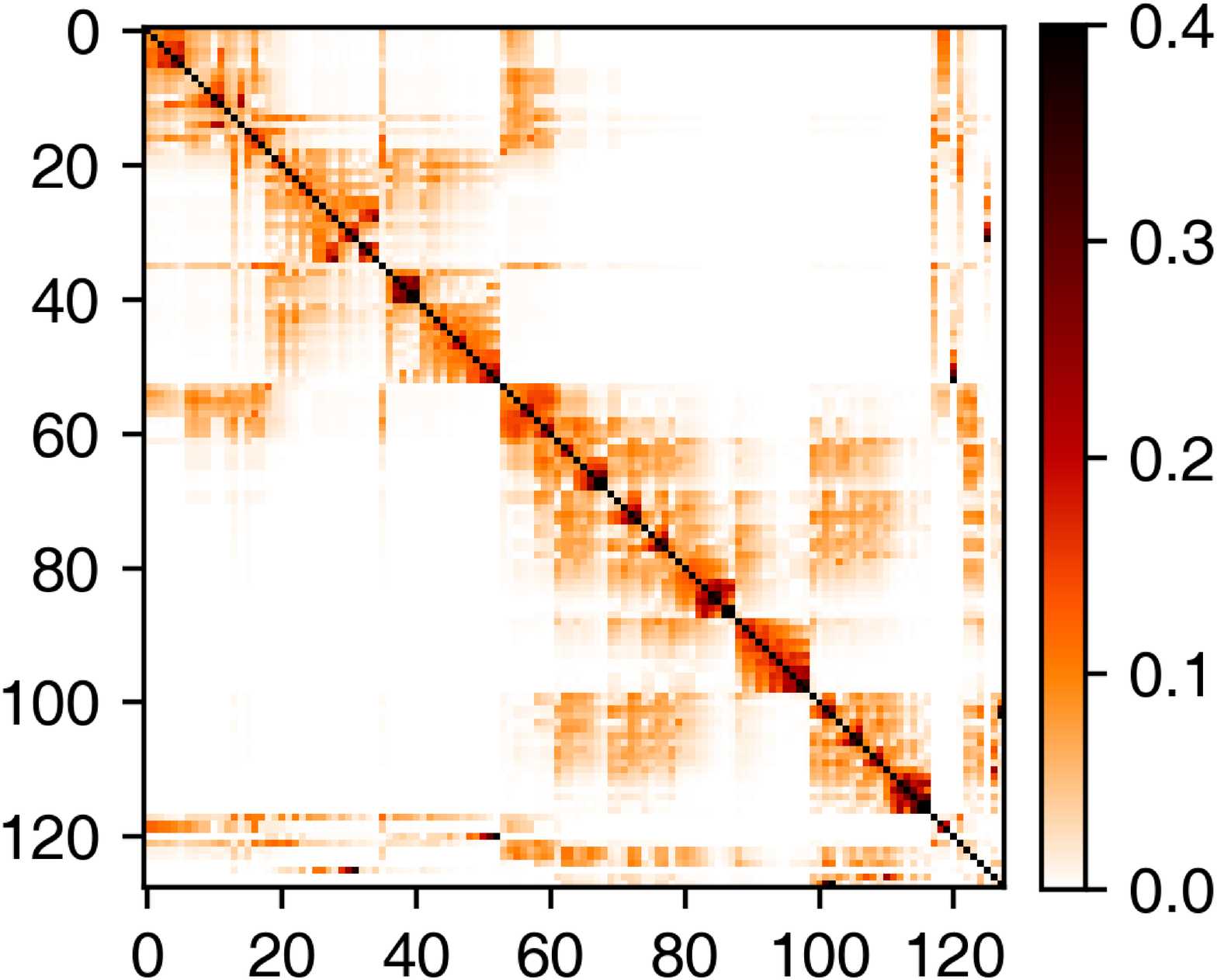}
}
\quad
\subfigure[]{
\includegraphics[width=3.7cm]{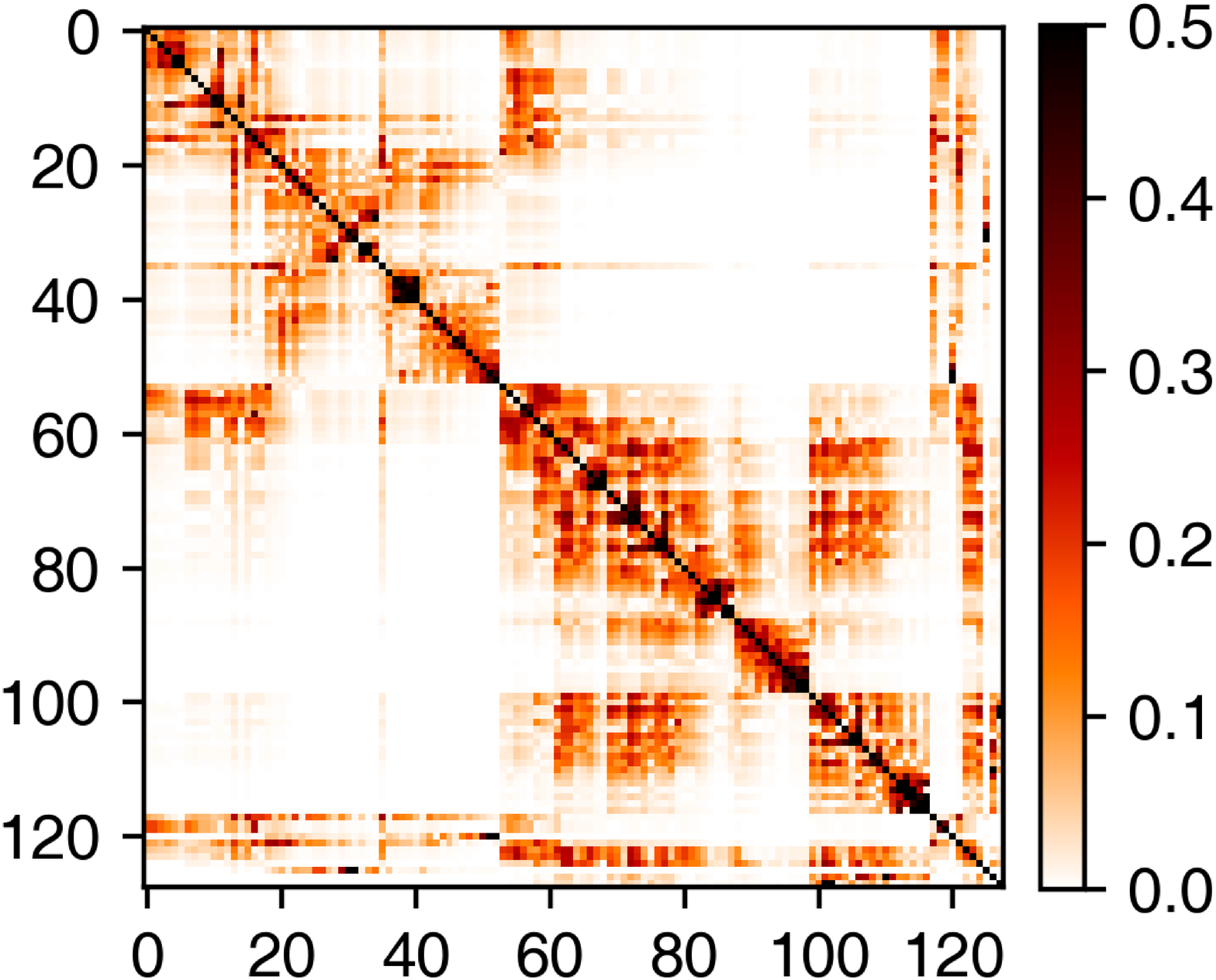}
}
\caption{Visualization of $\mathbf{\Delta}^m$ when $K_n=20$, (a) $m=1$, (b) $m=3$, (c) $m=5$, and (d) $m=10$. The indexes of the buses are sorted according to the order of bus numbers shown in Fig. \ref{IEEE123}. Absolute values of the entries in $\mathbf{\Delta}^m$ are visualized. Entries with values greater than certain thresholds (i.e., 0.2, 0.3, 0.4, and 0.5 for (a), (b), (c), and (d), respectively) have the same color.}
\label{vis2}
\end{figure}

We first visualize $\mathbf{\Delta}^m$ with different values of $m$ to illustrate the locality of the spectral filters, the results of which are shown in Fig. \ref{vis1} and Fig. \ref{vis2}. In Fig. \ref{vis1}, we illustrate the support of a filter when $m$ ranges from 1 to 4 (when $m=5$, the support of filters becomes the whole graph). In Fig. \ref{vis1}, the absolute values of the entries in $\mathbf{\Delta}^m$ are visualized. Although the size of filters grows fast with the increase of $m$, we can observe in Fig. \ref{vis2} that relatively large absolute values in $\mathbf{\Delta}^m$ are mainly limited to entries corresponding to bus pairs that are close to each other. Since the filters can be represented as polynomials of $\mathbf{\Delta}$, we conclude that the locality of filters are ensured when the value of $K_n$ is chosen properly. Note that higher-order terms in the polynomials facilitate the filters to explore more nodes in the graph.   

Three baseline models are also implemented for comparison:

\begin{enumerate}
\item SVM: The dimensionality of the measurements is reduced to 200 by principal component analysis (PCA). The radial basis function (RBF) kernel is used for the SVM with $\gamma = 0.002$ and $C = 1.5 \times 10^6$. LibSVM \cite{chang2011libsvm} in Python is used for the implementation in this paper.
\item Random forest (RF): The dimensionality of the measurements is also reduced to 200 by PCA. The number of trees is set to 300, the minimal number of samples per leaf is 1, while the minimal number of samples required for a split is set to 3.
\item Fully-connected neural network (FCNN): A three-layer FCNN is implemented as a vanilla baseline of neural networks. The numbers of hidden neurons for the three layers are 256, 128, and 64, respectively. Scaled exponential linear unit (SELU) is used as the activation function.
\end{enumerate} 
The hyper-parameters for SVM and RF are determined by 5-fold cross-validation. For the FCNN model, 10$\%$ of the training data is used to validate the hyper-parameters.

In order to justify the effectiveness of our proposed approach in real-world conditions including noise, measurement errors and communication errors, we add noise and errors to the measurements and compare the performance of different models. More specifically, three types of modifications of measurements are added:

\begin{enumerate}
\item Gaussian noise: We add Gaussian noises to the data so that the signal to noise rate (SNR) is 45 dB, as introduced in \cite{brown2016characterizing}. The noise has zero mean and the standard deviation, $\sigma_{\rm{noise}}$, is calculated as $\sigma_{\rm{noise}}=10^{-\frac{\text{SNR}}{20}}$.
\item Data loss of buses: We randomly drop the data of $N_{\rm{drop}}$ buses (i.e., set the measured values to 0) per data sample in the test dataset.
\item Random data loss for measured data: Each measurement at all buses is replaced by 0 with a probability $P_{\rm{loss}}$.
\end{enumerate} 
More specifically, we set $\sigma_{\rm{noise}}=10^{-\frac{45}{20}}$, $N_{\rm{drop}}=1$ and $P_{\rm{loss}}=0.01$ throughout the experiments unless otherwise specified. The detailed performance comparisons are given in the ensuing subsections.

\subsection{Fault Location Performance of the Models}

The fault location accuracies of various approaches are presented in Table \ref{Result-null}. In addition to the traditionally defined classification accuracy, we also use one-hop accuracy as a metric to measure the performance of the models. Specifically, a sample is considered correctly classified if the predicted faulty bus is directly connected to the actual faulty bus. For the GCN model, we repeat the trials three times and report the mean of the accuracy values.

In Table \ref{Result-null}, it is shown that the GCN model has the highest classification accuracy. SVM and RF (both with PCA) also have good performance, especially for one-hop accuracy. The accuracy obtained by FCNN is relatively low, but its one-hop accuracy is still satisfactory. 

\begin{table}[!t]
\renewcommand\arraystretch{1}
\centering  % 表居中
\captionsetup{justification=centering}
\caption{Fault Location Accuracies of Different Approaches} 
\begin{tabular}{p{2cm} p{1.4cm}<{\centering} p{2.3cm}<{\centering}}
\toprule[1.5pt]
Model & Accuracy & One-hop Accuracy\\
\midrule[0.75pt]
PCA + SVM                       & $94.60$          &  $98.31$    \\
PCA + RF                        & $94.96$          &  $99.28$    \\
FCNN                            & $84.64$          &  $96.38$    \\
GCN                             & $\bf{99.26}$    &   $\bf{99.93}$     \\
\bottomrule[1.5pt]
\end{tabular}
\label{Result-null}
\end{table}

The performance of the models with measurement modifications on the test dataset are shown in Table \ref{Result-mod}. Results corresponding to the individual and combined modifications are reported therein. A major observation is that the two data loss errors greatly lower the classification accuracy of the models. Nevertheless, the GCN model is quite robust to various modifications and significantly outperform other schemes. In addition, the FCNN model has higher accuracy than SVM and RF when data loss errors are involved, even though its classification accuracy is roughly 10\% lower than those two models.

\begin{table*}[!t]
\renewcommand\arraystretch{1}
\centering  % 表居中
\captionsetup{justification=centering}
\caption{Fault Location Accuracies of the Models Under Various Measurement Modifications} 
\begin{tabular}{P{1.5cm} P{1.8cm} P{1.8cm} P{1.8cm} P{1.8cm} P{1.8cm} P{1.8cm} P{1.8cm}}
\toprule[1.5pt]
Model & Noise (I) & Bus (II) & Random (III) & I + II & I + III & II + III & I + II + III \\
\midrule[0.75pt]
PCA + SVM                       & $89.13 \ /\ 97.30$   & $58.73\ /\ 79.97$ & $61.43\ /\ 81.78$   & $57.76\ /\ 79.26$ & $60.33\ /\ 81.09$ & $45.44\ /\ 69.64$ & $44.87\ /\ 69.20$   \\
PCA + RF                        & $85.94\ /\ 96.77$   & $53.82\ /\ 67.62$ & $58.57\ /\ 74.07$   & $52.15\ /\ 66.66$ & $56.94\ /\ 73.23$ & $40.55\ /\ 55.84$ &  $40.05\ /\ 55.64$  \\
FCNN                            & $85.72\ /\ 95.95$   & $62.61\ /\ 82.93$ & $69.40\ /\ 88.09$   & $61.24\ /\ 82.47$ & $69.51\ /\ 87.96$ & $53.54\ /\ 76.42$ & $54.12\ /\ 76.83$ \\
GCN                             & $\bf{97.10\ /\ 99.72}$ & $\bf{92.67\ /\ 97.44}$ & $\bf{89.09\ /\ 96.67}$ & $\bf{90.76\ /\ 97.83}$ & $\bf{87.70\ /\ 96.31}$ & $\bf{83.55\ /\ 94.51}$ & $\bf{80.63\ /\ 93.86}$      \\
\bottomrule[1.5pt]
\end{tabular}
\label{Result-mod}
\end{table*}

\begin{table*}[!t]
\renewcommand\arraystretch{1}
\centering  % 表居中
\captionsetup{justification=centering}
\caption{Fault Location Accuracies of the Models Under Various Measurement Modifications When Trained With Noisy Data} 
\begin{tabular}{P{1.5cm} P{2.1cm} P{2.1cm} P{2.1cm} P{2.1cm}}
\toprule[1.5pt]
Model & Noise & Noise + Bus &  Noise + Random & All Combined\\
\midrule[0.75pt]
PCA + SVM                       & $85.70\ /\ 96.21$       & $55.98\ /\ 77.74$                & $58.00\ /\ 80.17$    & $44.12\ /\ 68.51$ \\
PCA + RF                        & $86.51\ /\ 97.55$       & $64.11\ /\ 81.61$               & $66.12\ /\ 84.90$    & $52.34\ /\ 72.13$  \\
FCNN                            & $86.95\ /\ 97.19$       & $61.95\ /\ 82.58$              & $70.32\ /\ 88.52$   & $53.98\ /\ 76.55$  \\
GCN                             & $\bf{97.52\ /\ 99.73}$ & $\bf{92.67\ /\ 98.26}$         & $\bf{88.76\ /\ 96.44}$  &   $\bf{84.53\ /\ 94.77}$   \\
\bottomrule[1.5pt]
\end{tabular}
\label{Result-noise-mod}
\end{table*}

A more realistic setting is adding Gaussian noise to the data samples in the training dataset and observe the performance of the models. Table \ref{Result-noise-mod} gives the results of fault location accuracy corresponding to such a setup. The results for SVM and FCNN are in general consistent with the accuracy values in Table \ref{Result-mod}. For RF, however, the accuracies for modifications including data loss errors all increase by more than 10\%. Mild improvements are also observed for GCN. In summary, the GCN model has superior performance when measurement modifications are added to the data. Note that the modifications with data loss errors are not taken into account in the training phase. This indicates that the robustness of the GCN model may be generalizable to other types of errors in the data. In all subsequent experiments in the paper, the samples in both training and test datasets are added with Gaussian noise of 45 dB unless otherwise stated.

In the next subsection, we visualize the data upon the transformation by the FCNN model and the GCN model. Such visualizations facilitate our understanding of the performance differences induced by various schemes.

\subsection{Visualization of Data After Transformations}

Visualizing transformed data in two-dimensional spaces enables assessment of the ability of the models to extract useful information from the input data. In this paper, we use t-distributed stochastic neighbor embedding (t-SNE) with two components to visualize high-dimensional data \cite{maaten2008visualizing}. Specifically, t-SNE is used to investigate the local structure of the input data (i.e., normalized raw measurements), the data transformed by FCNN, and the data transformed by GCN. In particular, we are interested in studying how closely the samples corresponding to the same faulty bus are distributed.

\begin{figure}[!t]
\setlength{\abovecaptionskip}{0pt}
\centering
\includegraphics[width=5cm]{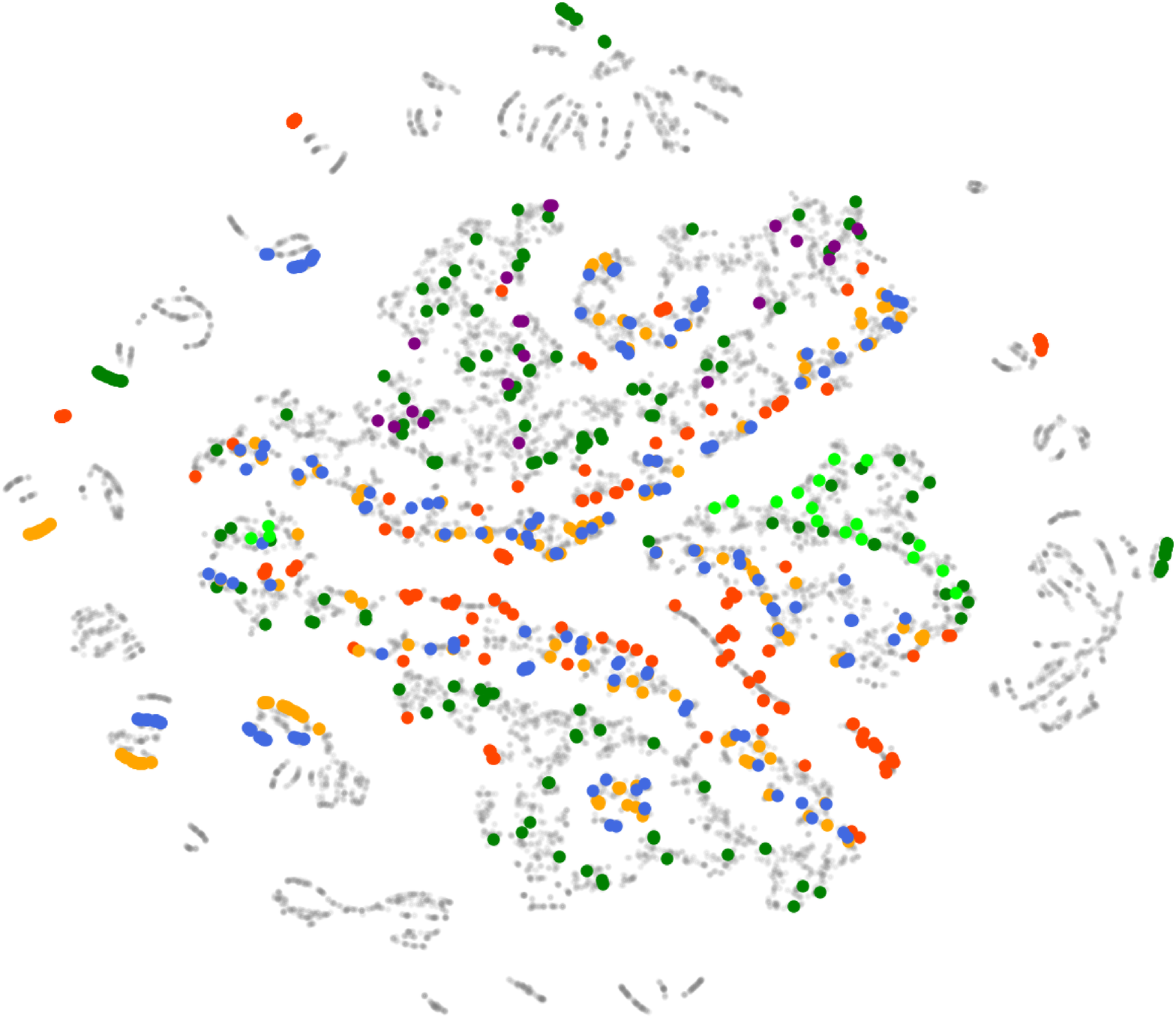}
\caption{Visualization of test data processed by PCA with 200 components and t-SNE with two components. Dots with the same color (except for small gray dots) correspond to the same faulty bus.}
\label{scatter_ori}
\end{figure}

In Fig. \ref{scatter_ori}, we visualize the data samples in the test dataset with t-SNE after the dimensionality of data is reduced to 200 by PCA, which is also used to speed up the calculation process of t-SNE). In order to highlight the distribution of data belonging to the same class (faulty bus), 6 groups of data samples of bus 1, 21, 66, 85, 111, and 250 are marked with colors. Data samples of other buses are plotted as the gray dots. It can be seen in the figure that the dots of different colors scatter around such that it is hard to separate the data samples from different classes.

\begin{figure}[!t]
\setlength{\abovecaptionskip}{0pt}
\centering
\subfigure[]{                   
\begin{minipage}{4cm}
\centering   
\includegraphics[width=4cm]{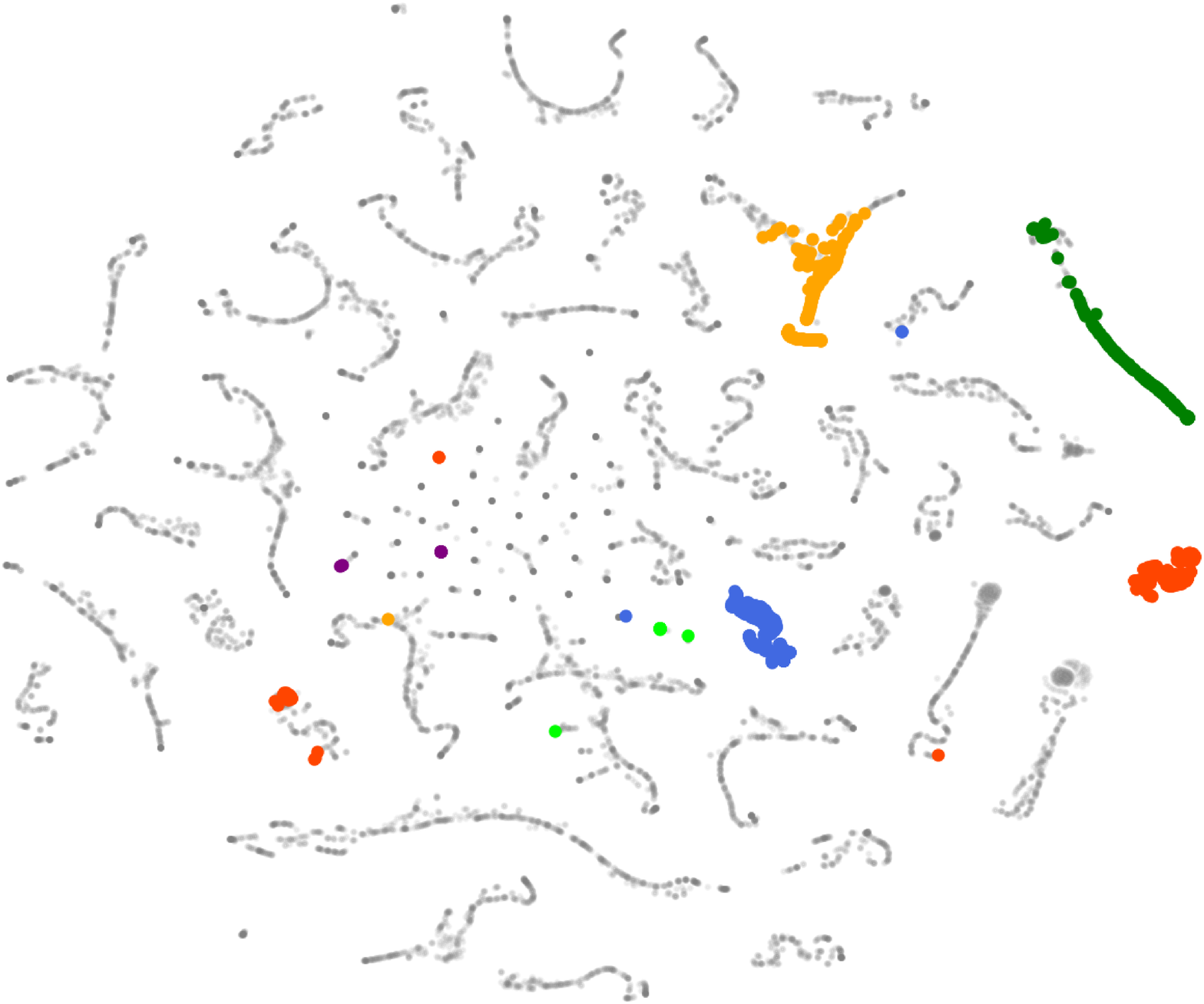}
\end{minipage}
}
\subfigure[]{                
\begin{minipage}{4cm}
\centering  
\includegraphics[width=4cm]{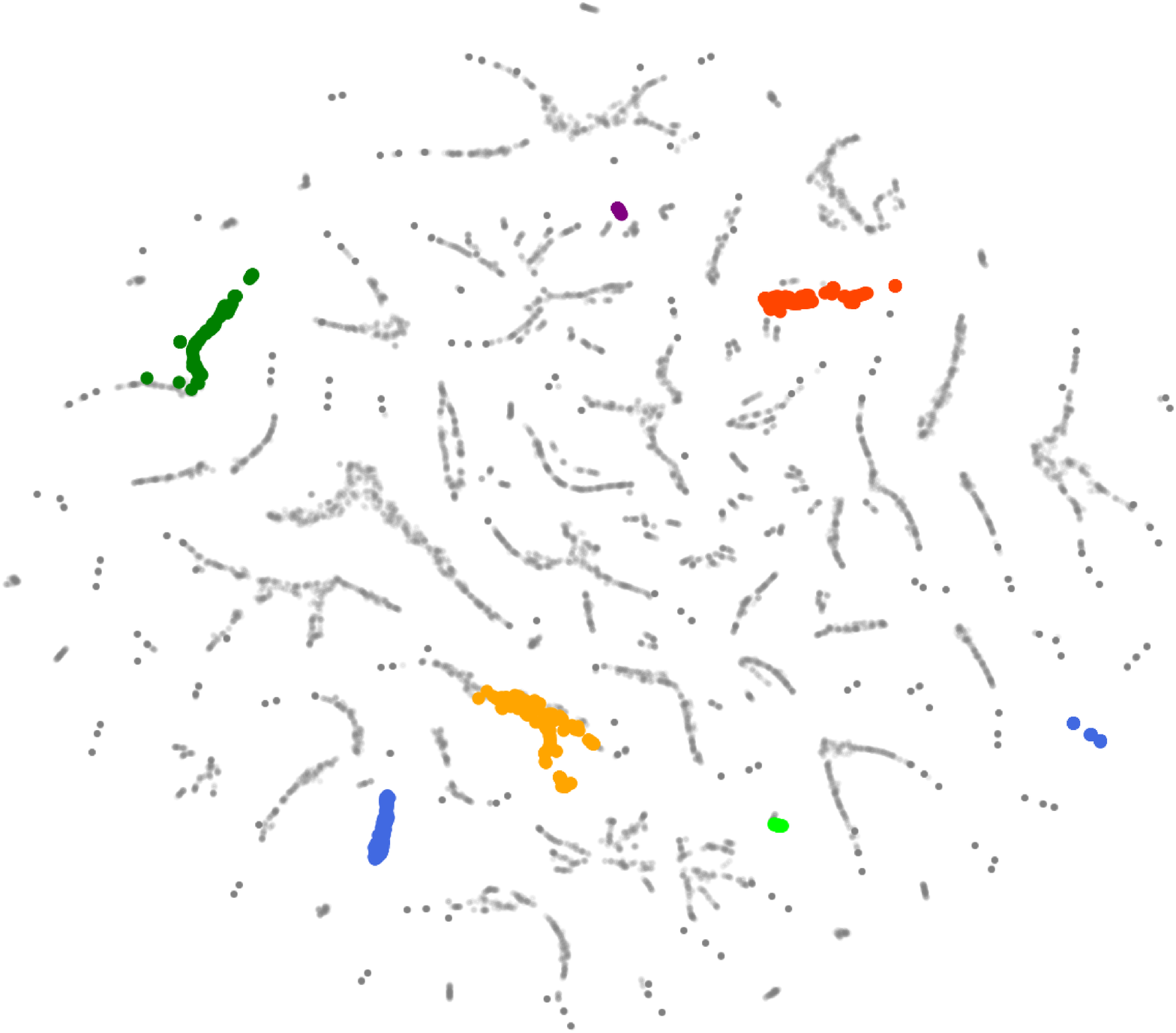}
\end{minipage}
}
\caption{Visualization of hidden features of test data added with Gaussian noise using t-SNE with two components: (a) the FCNN model, and (b) the GCN model. The models are trained with added Gaussian noise.}
\label{scatter_noise}
\end{figure}

We then visualize the data samples in the test dataset after they are transformed by the FCNN and the GCN models, as shown in Fig. \ref{scatter_noise}. Both models are trained with added Gaussian noise while the test data is also added with Gaussian noise. We extract the data from the outputs of the fully-connected layer right before the final output layer. For the FCNN model, each data sample is 64-dimensional, while the dimensionality of data samples is 256 for the GCN model. In Fig. \ref{scatter_noise} (a), the data samples of the same class hardly cluster together, except for the dark green dots in the upper-right corner. In Fig. \ref{scatter_noise} (b), however, most samples of the same color appear closely together, except that only a small fraction of blue dots are separated from its main cluster. Note that the visualization in Fig. \ref{scatter_noise} corresponds to the ``Noise'' column of Table \ref{Result-noise-mod}. That is, the improved feature extraction capability of the GCN model gives a performance boost in classification accuracy of more than 10\%.

\begin{figure}[!t]
\setlength{\abovecaptionskip}{0pt}
\centering
\subfigure[]{                   
\begin{minipage}{4cm}
\centering   
\includegraphics[width=4cm]{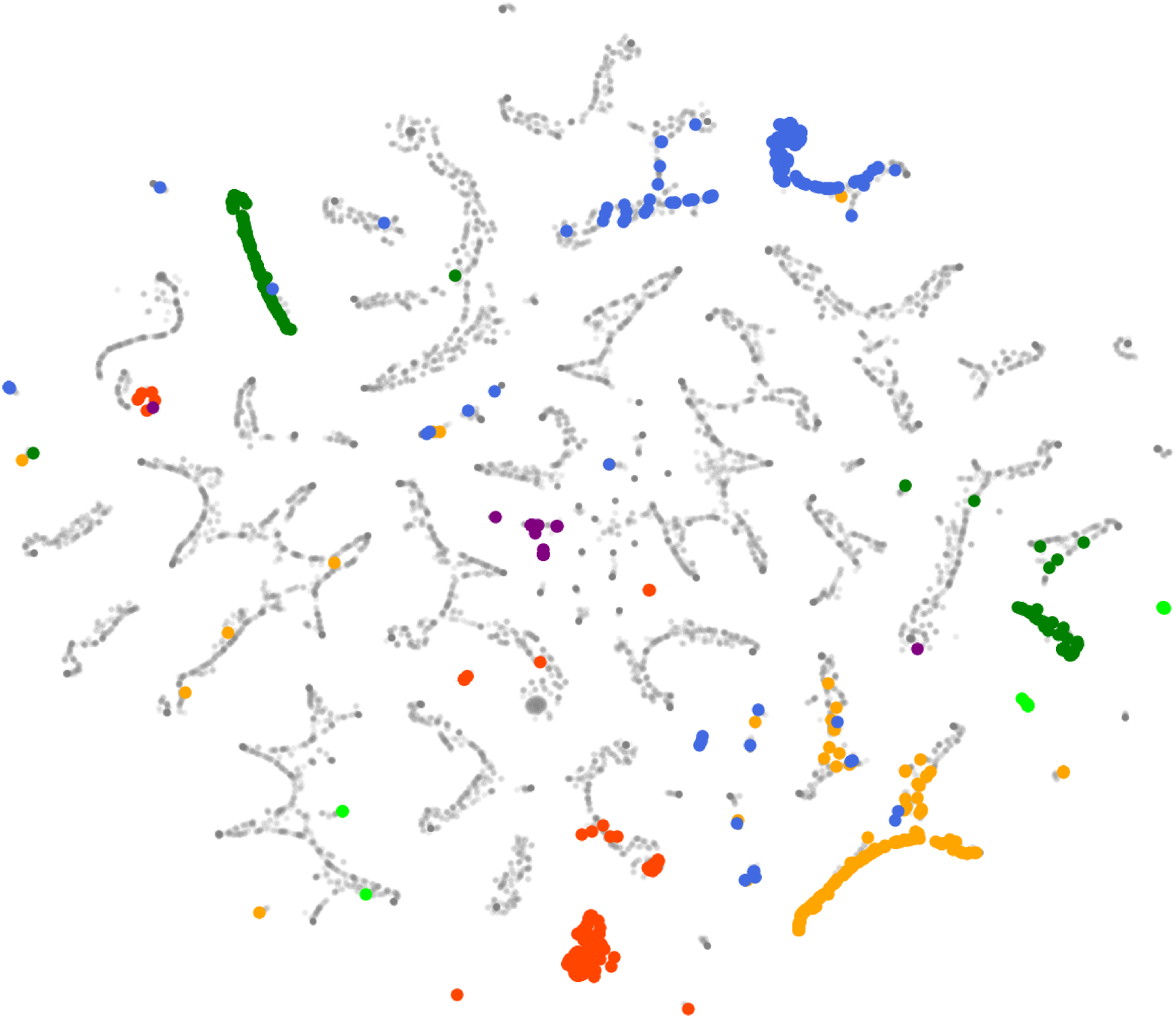}
\end{minipage}
}
\subfigure[]{                
\begin{minipage}{4cm}
\centering  
\includegraphics[width=4cm]{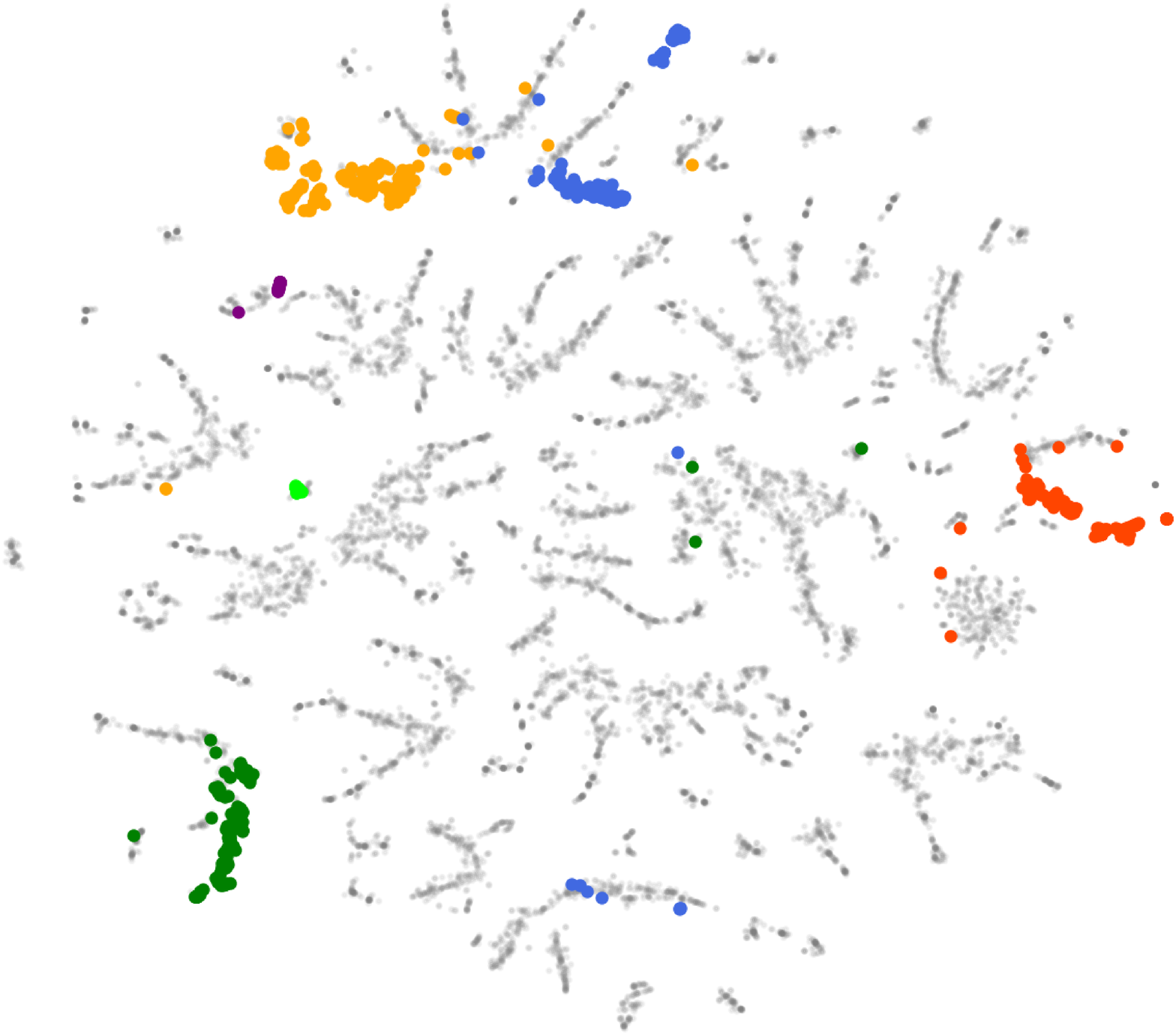}
\end{minipage}
}
\caption{Visualization of hidden features of test data added with all three types of modifications using t-SNE with two components: (a) the FCNN model, and (b) the GCN model. The models are trained with added Gaussian noise.}
\label{scatter_combined}
\end{figure}

As shown in Table \ref{Result-mod} and Table \ref{Result-noise-mod}, the two types of data loss errors have significant impact on the classification performance of all models. Thus, in Fig. \ref{scatter_combined} we proceed to visualize the data samples that are added with Gaussian noise and two types of data loss errors. A lot of small sample clusters of the six colored faulty buses can be seen at multiple locations in Fig. \ref{scatter_combined} (a), which indicates that the FCNN model has difficulty in generalizing its feature extraction capability to the data modified with the two types of data loss errors. On the contrary, the GCN model still preserves the structures of the data to a large extend. The proportion of data samples that are separated from the main clusters is relatively small. Such a capability of preserving data structure gives rise to more than 30\% performance gain for the proposed GCN, as shown in the last column of Table \ref{Result-noise-mod}.

\subsection{Increasing Model's Robustness by Data Augmentation}

We have shown that the GCN model is quite robust to mild noise and data loss errors (i.e., the SNR is 45 db, $N_{\rm{drop}}=1$, and $P_{\rm{loss}}=0.01$). The data collected from the field, however, may have lower SNR and higher data loss rates. Thus, it is desirable if the model is able to generalize to different levels of noise and data loss errors. In light of this, we implement data augmentation during training of the model by adding various levels of noise and data loss errors to the input data. Specifically, for the $i$th input sample in a mini-batch, we first add Gaussian noises with $\sigma_{\rm{noise}} = \tilde{\sigma}_i$ to the measurements and randomly set measurements to 0 with $N_{\rm{drop}}=\tilde{n}_i$ and $P_{\rm{loss}}=\tilde{p}_i$. We randomly choose $\tilde{\sigma}_i$, $\tilde{n}_i$, and $\tilde{p}_i$ from [0, $10^{-\frac{45}{20}}$, $10^{-\frac{40}{20}}$, $10^{-\frac{35}{20}}$, $10^{-\frac{30}{20}}$, $10^{-\frac{25}{20}}$], [0, 1, 2, 3, 4, 5], and [0, 0.01, 0.02, 0.03, 0.04, 0.05], respectively, with equal probability. Note that the data augmentation is applied to each mini-batch, thus a new data sample is generated for the $j$th data sample in each epoch unless $\tilde{\sigma}_j=\tilde{n}_j=\tilde{p}_j=0$, in which case the data sample is unchanged.

\begin{figure}[!t]
\setlength{\abovecaptionskip}{0pt}
\centering
\includegraphics[width=7cm]{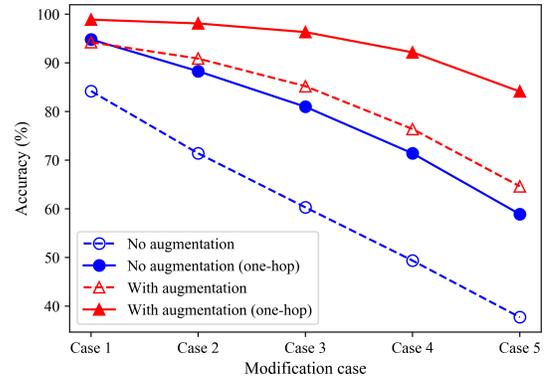}
\caption{Performance of the proposed GCN model with and without data augmentation. The values of $\sigma_{\rm{noise}}$, $N_{\rm{drop}}$, and $P_{\rm{loss}}$ are ($10^{-\frac{45}{20}}$, 1, 0.01), ($10^{-\frac{40}{20}}$, 2, 0.02), ($10^{-\frac{35}{20}}$, 3, 0.03), ($10^{-\frac{30}{20}}$, 4, 0.04), and ($10^{-\frac{25}{20}}$, 5, 0.05) for the 5 cases.}
\label{augmentation}
\end{figure}

We report the performance of the GCN model under various noise and data loss levels with and without data augmentation in Fig. \ref{augmentation}. Specifically, case 1 has the lowest level of noise and data loss errors while case 5 has the highest level. It is shown in the figure that the proposed data augmentation procedure greatly improves the fault location accuracies, and the one-hop accuracies for cases 1, 2, and 3 are higher than 95\%. The one-hop accuracy when data augmentation is applied is higher than 84\% even for case 5, for which the SNR is 25, measurements from 5 buses are lost, and each measurement may also be lost with a probability of 0.05. 

Although the accuracy with data augmentation for case 1 is quite high, some samples are still assigned to wrong buses. In order to examine the characteristics of misclassified samples, we collect the samples with predicted faulty buses more than two hops away from the correct faulty buses. Note that random noise and data losses are added to the test samples, so the results are different for each trial. Specifically, the collected bus pairs are (25r, 250), (30, 25r), (33, 25r), (53, 56), (61s, 62), (61s, 68), (89, 79), (92, 95), (99, 105), (108, 102), (151, 49), (250, 28), and (250, 25r) (the first number is the correct bus). The majority of the two buses in the bus pairs are three hops away. In addition, 4 buses, namely, 29, 34, 76, and 108 are used to illustrate the characteristics of misclassified samples for case 5. The buses are located near the four corners of the network illustrated in Fig. \ref{IEEE123}. While bus 34 has no misclassified samples, the lists of wrong predictions more than two hops away for bus 29, 76, and 108 are [18 (5), 21 (4), 22 (5), 23 (3), 26 (3), 27 (4), 31 (4)], [60 (3), 61s (4), 66 (8), 74 (3), 75 (4), 79 (3), 81 (4), 83 (6), 97 (3), 99 (5)], and [61s (6), 97 (3), 98 (4), 100 (6), 111 (3), 450 (7)], respectively (the number of hops between the correct and predicted buses are included in the brackets). While hop numbers up to 8 exist in the results, most of the hop numbers are lower than 5. As the two-hop accuracy for case 5 is 92.49\%, it can be concluded that the GCN model is able to locate a fault within the vicinity of its exact location under severe data loss errors in most cases.

\begin{figure}[!t]
\setlength{\abovecaptionskip}{0pt}
\centering
\includegraphics[width=7cm]{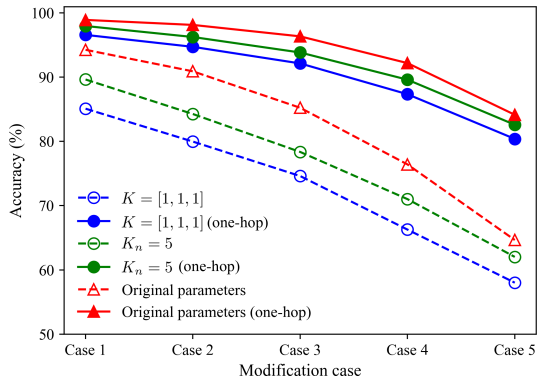}
\caption{Performance of the proposed GCN model with data augmentation and different hyper-parameter choices. The original hyper-parameters are $K=[3,4,5]$ and $K_n = 20$. Two experiments, namely, changing $K$ to [1,1,1] and changing $K_n$ to 5 are implemented for comparison.}
\label{augmentation_reception}
\end{figure}

In previous experiments, the values for the hyper-parameters $K_n$ and $K$ are 20 and [3, 4, 5] (the $K$ for 3 graph convolution layers), respectively. It is expected that when data losses occur in some of the measurements close to a fault, information from other measurements may help the model locate the fault. In order to justify the choice of $K_n$ and $K$, we report the performance of the GCN model with relatively small values of $K_n$ and $K$ in Fig. \ref{augmentation_reception}. Specifically, for one experiment we change $K_n$ to 5, and for the other experiment we set $K$ to [1, 1, 1]. It can be observed that the GCN model with the original hyper-parameters has higher accuracies. In addition, setting $K$ to [1, 1, 1] hurts the performance of the model as it severely limits the range of measurements a node in the GCN can obtain information from.

\subsection{Performance Under Distribution Network Reconfiguration}

The configuration of a distribution network may change in order to reduce loss or balance the loads \cite{baran1989network}. The reconfiguration of a network can be achieved by opening some of the normally closed switches and closing some of the normally open switches. The IEEE 123 bus system has a three-phase normally open switch between node 151 and node 300 (see Fig. \ref{IEEE123}), which can be used to guarantee electricity supply of the system if some of the normally closed switches are open.

In this work, we consider two cases of network reconfiguration:
\begin{enumerate}
\item Open the switch connecting node 18 and node 135, and close the switch connecting node 151 and node 300.
\item Open the switch connecting node 97 and node 197, and close the switch connecting node 151 and node 300.
\end{enumerate}
In order to evaluate the performance of the proposed model under network reconfiguration, we generate 5 samples for each fault type at each bus for the two cases and directly use the GCN model trained with data augmentation to identify the faulty buses. As a result, the fault location accuracies and one-hop accuracies for the two cases are 88.37\% (98.28\%), and 91.89\% (98.61\%), respectively. As the reconfiguration scenarios are not considered during the training phase of the model, the results indicate that the GCN model has high stability against unseen network reconfiguration scenarios.

\subsection{Performance Under Multiple Connection Scenarios of Branches}

In this subsection, the performance of the models under multiple connection scenarios of several branches is examined. Specifically, the connected phase of a branch in a distribution network may change from time to time and it is expected that the model can deal with such changes. Three branches in the IEEE 123 bus system are considered: 
\begin{enumerate}
\item The branch connecting bus 36, 38, and 39.
\item The branch connecting bus 67 to 71.
\item The branch connecting bus 108 to 114. 
\end{enumerate}
Phase 1 and 2 of bus 36, and all three phases of bus 67 and bus 108 are connected to the distribution system. The buses on the branches use only one of the phases for connection.

Unlike network reconfiguration achieved by opening and closing of switches, changing the connected phase of a single-phase branch requires additional data to train the models. We implement a simple data generation process in order to add data with changed phases of aforementioned branches into the training and test datasets. Specifically, we change the phase of only one of the branches to another available phase and generate 5 data samples for each fault type at each bus. Thus, both the new training and test datasets contain 30420 data samples.

Fault location accuracies of the models with additional phase-changed data are presented in Table \ref{Result-pc}. The results for faults at all buses and at modified buses with changed phases are included. For all schemes, the accuracies for faults at modified buses are lower than the counterparts of faults at all buses. The GCN model has the highest accuracies for all scenarios while the one-hop accuracies are more than 99\%. Comparing the column of ``All Buses'' of Table \ref{Result-pc} with the ``Noise'' column of Table \ref{Result-noise-mod}, we can see that the additional data has almost no impact on GCN, while the accuracies for other models decrease by 1-5\%. Thus, we conclude that the GCN model is robust to the change of connected phases of single branches if the training dataset covers samples of the additional connection scenarios.

\begin{table}[!t]
\renewcommand\arraystretch{1}
\centering  % 表居中
\captionsetup{justification=centering}
\caption{Fault Location Accuracies of the Models with Additional Data Generated With Changed Phases of Chosen Branches} 
\begin{tabular}{p{2cm} p{2cm} p{2cm}}
\toprule[1.5pt]
Model & All Buses & Modified Buses \\
\midrule[0.75pt]
PCA + SVM                       & $81.87\ /\ 94.09$          &  $77.47\ /\ 92.00$    \\
PCA + RF                        & $81.57\ /\ 95.11$          &  $79.64\ /\ 93.93$    \\
FCNN                            & $85.38\ /\ 96.20$          &  $82.85\ /\ 95.36$    \\
GCN                             & $\bf{97.65\ /\ 99.77}$    &   $\bf{93.66\ /\ 99.38}$     \\
\bottomrule[1.5pt]
\end{tabular}
\label{Result-pc}
\end{table}

\subsection{Performance on High Impedance Faults}

The detection of high impedance faults in distribution networks is a challenging task as the current magnitude is generally close to the level of load current \cite{ghaderi2017high}. In this subsection, we evaluate the performance of the GCN model on locating high impedance faults. Specifically, we add single-phase-to-ground faults with high fault resistance to the training dataset and report the fault location results on various ranges of fault resistance. For the construction of the training dataset, in addition to the data samples with small fault resistance values, we generate 40 samples for each phase at each bus and the fault resistance is uniformly sampled between 100 $\Omega$ and 5000 $\Omega$. Five fault resistance ranges, namely, 100 $\Omega$ to 1000 $\Omega$, 1000 $\Omega$ to 2000 $\Omega$, 2000 $\Omega$ to 3000 $\Omega$, 3000 $\Omega$ to 4000 $\Omega$, and 4000 $\Omega$ to 5000 $\Omega$ are used to generate test samples. For each range, 5 test samples are generated for each type of fault at each bus. In order to test the generalizability of the GCN model, we further split the fault resistance ranges into two sets of intervals (the length of each interval is 10 $\Omega$), namely, $\{ R_1 | 20k < R_1 < 20k + 10, k \in \mathbb{Z} : 5 \leq k \leq 249 \}$, and $\{ R_2 | 20k + 10 < R_2 < 20(k + 1), k \in \mathbb{Z} : 5 \leq k \leq 249 \}$, where $R_1$ is used for samples in the training dataset and $R_2$ is used to generate samples in the test datasets. 

\begin{figure}[!t]
\setlength{\abovecaptionskip}{0pt}
\centering
\includegraphics[width=7cm]{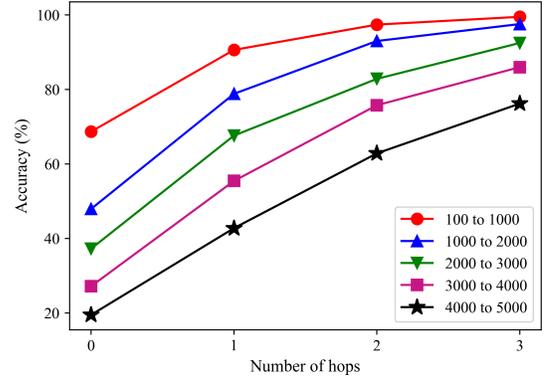}
\caption{Performance of the proposed GCN model for high impedance faults within various fault resistance ranges.}
\label{high_impedance}
\end{figure}

The accuracies for high impedance faults with different ranges of fault resistance are shown in Fig. \ref{high_impedance}. As the zero-hop and one-hop accuracies are relatively low, we also report the two-hop and three-hop accuracies. It is seen in the figure that with the increase of fault resistance, the fault location accuracy drops. Although the increased fault resistance makes it hard to find the exact fault location, the accuracies increase rapidly with the increase of the number of hops, which indicates that the model is still able to capture a part of fault characteristics for high impedance faults.  

\subsection{Discussion on the Types of Measurements Used for the Model}

As the proposed GCN model uses amplitudes and phase angles of both voltage and current measurements as inputs, it is necessary to examine the contributions of the different measurements to the performance of the model. Specifically, the first concern is that the measured currents are the injected currents at the loads, which provide less information about the faults compared with currents flowing in the branches connecting the buses. The second concern is the contribution of phase angle to the fault location accuracy as measuring the phase angle requires additional installation of phasor measurement devices.

\begin{table*}[!t]
\renewcommand\arraystretch{1}
\centering  % 表居中
\captionsetup{justification=centering}
\caption{Fault Location Accuracies of the GCN Model With Different Measurement Scenarios} 
\begin{tabular}{P{3.5cm} P{2.1cm} P{2.1cm} P{2.1cm} P{2.1cm}}
\toprule[1.5pt]
Measurement Scenario & Noise & Noise + Bus &  Noise + Random & All Combined\\
\midrule[0.75pt]
Voltage amplitudes  & $94.66\ /\ 99.36$       & $67.51\ /\ 82.91$       & $75.55\ /\ 87.38$    & $58.02\ /\ 75.34$  \\
Voltage phasors     & $97.43\ /\ 99.88$       & $90.38\ /\ 97.49$       & $89.82\ /\ 97.33$   & $83.54\ /\ 94.63$  \\
Current phasors     & $91.04\ /\ 98.43$       & $82.40\ /\ 93.64$       & $83.04\ /\ 94.44$   & $76.32\ /\ 90.44$  \\
Voltage and current phasors   & $97.52\ /\ 99.73$ & $92.67\ /\ 98.26$  & $88.76\ /\ 96.44$  &   $84.53\ /\ 94.77$   \\
\bottomrule[1.5pt]
\end{tabular}
\label{Result-measurement}
\end{table*}

In light of the concerns, we compare the performance of the GCN model under different measurement scenarios and present the results in Table \ref{Result-measurement}. The results for the last row are the same as the row of GCN in Table \ref{Result-noise-mod}. It is shown in the table that the results with voltage phasors are quite similar to the results with both voltage and current phasors. The results with current phasors, however, is much lower than the results with voltage phasors. Further, when only voltage amplitudes are used, the accuracies with data loss errors are dramatically lower than other scenarios. Two conclusions can be drawn from the results:
\begin{enumerate}
\item For the design of the GCN model in this paper, the performance of the model mainly relies on the voltage phasors. Other types of current measurements such as currents flowing in the branches may be added in order to improve the fault location accuracy.
\item It is important to include phase angles in the inputs for the GCN model, especially when data loss errors are considered.
\end{enumerate}

\subsection{Implementation of the GCN Model on Another Distribution Network}

\begin{figure}[!t]
\setlength{\abovecaptionskip}{0pt}
\centering
\includegraphics[width=6.5cm]{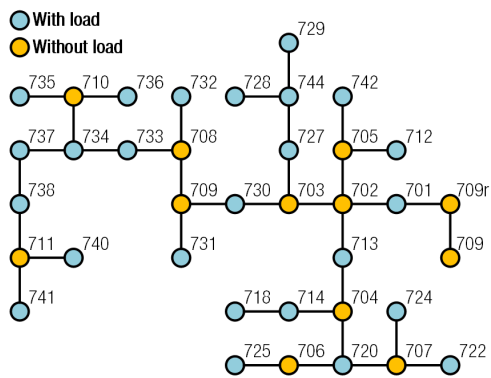}
\caption{An illustration of the IEEE 37 bus system.}
\label{IEEE37}
\end{figure}

As the previous experiments are all carried out in the IEEE 123 bus system, we implement the GCN model to the IEEE 37 bus system to verify that the model can perform well in a new distribution network. The topology of the IEEE 37 bus system is shown in Fig. \ref{IEEE37}. Similar to the implementation for the IEEE 123 bus system, we measure the voltage and current phasors at the phases connected to loads. The generation scheme for training and test datasets described in Section II. D is used. A series of hyper-parameters are used to see if the performance of the model is sensitive to the choice of hyper-parameters.

\begin{table}[!t]
\renewcommand\arraystretch{1}
\centering  % 表居中
\captionsetup{justification=centering}
\caption{Fault Location Accuracies of the GCN Model for the IEEE 37 Bus System With Different Values of $K_n$ and $K$} 
\begin{tabular}{p{3cm} p{1.1cm} p{1.1cm} p{1.1cm}}
\toprule[1.5pt]
Hyper-parameters & Zero-hop & One-hop & Two-hop \\
\midrule[0.75pt]
$K_n=10, K=[1,2,3]$       & $88.70$      & $\underline{95.31}$        & $ 96.94 $            \\
$K_n=10, K=[2,3,4]$       & $88.94$      & $96.36$        & $ 97.03 $            \\
$K_n=10, K=[3,4,5]$       & $\bf{89.97}$      & $\bold{97.14}$        & $ \bold{97.89} $          \\
$K_n=5, K=[2,3,4]$        & $88.90$      & $\underline{95.04}$        & $ 96.81 $       \\
$K_n=15, K=[2,3,4]$       & $\bf{90.05}$      & $\bold{96.93}$        & $ \bold{97.73} $       \\
$K_n=20, K=[2,3,4]$       & $89.51$      & $96.20$        & $ 97.25 $       \\
$K_n=5, K=[1,1,1]$        & $\underline{88.12}$      & $95.59$        & $ \underline{96.48} $       \\
$K_n=10, K=[1,1,1]$       & $\underline{88.18}$      & $\underline{95.31}$        & $ \underline{96.25} $       \\
\bottomrule[1.5pt]
\end{tabular}
\label{Result-37bus}
\end{table}

We first evaluate the performance of the GCN model with different values of $K_n$ and $K$ (the other hyper-parameters remain unchanged), and the accuracies are shown in Table \ref{Result-37bus}. For each column in the table, the lowest two values are highlighted with underlines while the highest two values are marked in bold. It is observed that the performance of the GCN model is quite stable under different values of $K_n$ and $K$. In Section III. D, we have shown that properly choosing the values of $K_n$ and $K$ can increase the model's robustness against data loss errors. When data loss errors are not considered, however, the negative effect of setting $K_n$ and $K$ to small values is insignificant. This indicates that laborious tuning of hyper-parameters is not needed when implementing the GCN model to a new distribution network. 

\begin{table}[!t]
\renewcommand\arraystretch{1}
\centering  % 表居中
\captionsetup{justification=centering}
\caption{Fault Location Accuracies of the GCN Model for the IEEE 37 Bus System With Different Numbers of Layers} 
\begin{tabular}{p{3.1cm} p{1.1cm} p{1.1cm} p{1.1cm}}
\toprule[1.5pt]
Hyper-parameters & Zero-hop & One-hop & Two-hop \\
\midrule[0.75pt]
$K=[2]$ (1 layer)       & $74.80$      & $88.55$        & $ 93.06 $            \\
$K=[2,3] $ (2 layers)       & $85.26$      & $94.37$        & $ 95.59 $            \\
$K=[2,3,4] $ (3 layers)       & $\bf{88.94}$      & $\bf{96.36}$        & $ \bf{97.03} $          \\
\bottomrule[1.5pt]
\end{tabular}
\label{Result-layer}
\end{table}

Another important hyper-parameter is the number of graph convolution layers. Although the GCN model does not require a large number of layers, it is expected that the model may not have enough learning capacity when the number of layers is not enough. We report the performance of the GCN model with different numbers of graph convolution layers in Table \ref{Result-layer}. It is clearly observed in the table that increasing the number of layers improves the fault location accuracy, but the gain of adding another layer decreases when the third layer is added.

\begin{figure}[!t]
\setlength{\abovecaptionskip}{0pt}
\centering
\includegraphics[width=7cm]{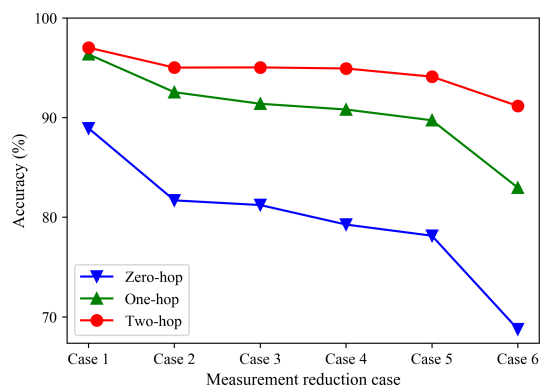}
\caption{Performance of the proposed GCN model with different cases of reduction on number of measured buses. Case 1 represents the original scenario with 25 measured buses.}
\label{number_measurements}
\end{figure}

Finally, we use the IEEE 37 bus system to discuss some practicability issues. The first concern is the number of measured buses required for the GCN model. In the original scenario, the phases connected to loads at 25 buses are monitored. We gradually reduce the number of monitored buses and compare the performance of the model under these reduction cases. Specifically, the lists of removed buses are [714, 733, 737, 738, 744], [720, 727, 734], [724, 728, 735], [701, 712, 713, 730, 741], and [718, 722, 729, 731, 736]. Although this is only one possibility of reduction, we try to remove the buses evenly across the network to avoid large areas of unmonitored buses. When the 4th list of buses are removed, only 9 buses at the end of branches are left. The final reduction case leaves the model with only 4 measured buses, namely, 725, 732, 740, and 742. The accuracies of the different reduction cases are illustrated in Fig. \ref{number_measurements}. Apparently, reducing the number of measured buses has a negative effect on the fault location accuracies. The two-hop accuracy, however, is not very sensitive to the reduction of measured buses until the last reduction. Four case 5, specifically, the two-hop accuracy is 94.12\% with measurements from only 9 buses at branch ends. The results indicate that it is harder for the GCN model to find the exact fault locations when a large proportion of the measured buses are excluded, but the ability to find the vicinity of the faulty bus only requires a small proportion of buses to be measured.

\begin{table}[!t]
\renewcommand\arraystretch{1}
\centering  % 表居中
\captionsetup{justification=centering}
\caption{Fault Location Accuracies of the GCN Model for the IEEE 37 Bus System With Different Sizes of Training Datasets} 
\begin{tabular}{p{1.5cm} p{1.1cm} p{1.1cm} p{1.1cm}}
\toprule[1.5pt]
Dataset Size & Zero-hop & One-hop & Two-hop \\
\midrule[0.75pt]
100\%       & $88.94$      & $96.36$        & $ 97.03 $            \\
50\%        & $66.99$      & $87.04$        & $ 93.73 $            \\
25\%        & $45.46$      & $73.33$        & $ 88.69 $          \\
10\%        & $21.42$      & $46.79$        & $ 67.49 $          \\
5\%         & $13.21$      & $33.27$        & $ 52.73 $          \\
\bottomrule[1.5pt]
\end{tabular}
\label{Result-size}
\end{table}

The second concern is the number of training data needed to train the GCN model. The above mentioned dataset for the IEEE 37 bus system contains 20 samples for each fault type at each bus. With this as the size of 100\%, we reduce the number of generated samples for each fault type at each bus to 10, 5, 2, and 1, and compare the accuracies of the scenarios in Table \ref{Result-size}. The size of the test dataset remain the same. It is seen in the table that the performance of the GCN model degrades as the size of the training dataset reduces. With only 1 sample for each fault type at each bus, the two-hop accuracy is a little above 50\%. As it is hard to collect field data with varied fault types, fault resistances and load levels, a more practical solution is to combine field data with synthetic data simulated according to the need of the model. With the help of transfer learning \cite{pan2009survey}, the model can transfer the knowledge learned from simulated data to locate faults using actual measurements from the system. Such a problem formulation is beyond the scope of this paper, but the results in our work provides an upper bound for the performance of the GCN model as we use simulated data only.  

\section{Conclusion and Future Work}

In this paper, we develop a GCN model for the task of fault location in distribution systems. Simulation results tested with the IEEE 123-bus and 37-bus systems show that the proposed GCN model is significantly effective in processing fault-related data. The proposed model is more robust to measurement errors compared with many other machine learning approaches including SVM, RF, and FCNN. Visualization of the activations of the last fully-connected layer shows that the GCN model extracts features that are robust to missing entries in the measurements. Further experiments show that the model can adapt to topology changes and perform well with a limited number of measured buses. In a nutshell, the present paper proposes a flexible and widely-applicable energy data analytics framework for improving situational awareness in power distribution systems.

The proposed framework and approach open up a few interesting research directions. First, the effectiveness of the GCN model in more realistic settings needs further investigation (e.g., use field data to fine-tune the model trained with synthetic data, or train the model with both field data and synthetic data by transfer learning). Second, it is valuable to develop new schemes for transferring a learned model to other distribution systems with different topologies. A new challenge comes from the integration of distributed generation, which introduces high-level uncertainties into the grids, and may alter the characteristics of the measurements during faults.

\section*{Acknowledgement}
The authors are grateful for the support of NVIDIA Corporation with the donation of the Titan Xp GPU used for this project. We would also like to thank Ron Levie at TU Berlin, and Federico Monti at University of Lugano, who helped us with the implementation of the GCN model. 

{
\small
\bibliographystyle{IEEEtran}%
\bibliography{GCN.bib}
}

\vspace{-25pt}

\begin{IEEEbiography} [{\includegraphics[width=1in,height=1.25in,clip,keepaspectratio]{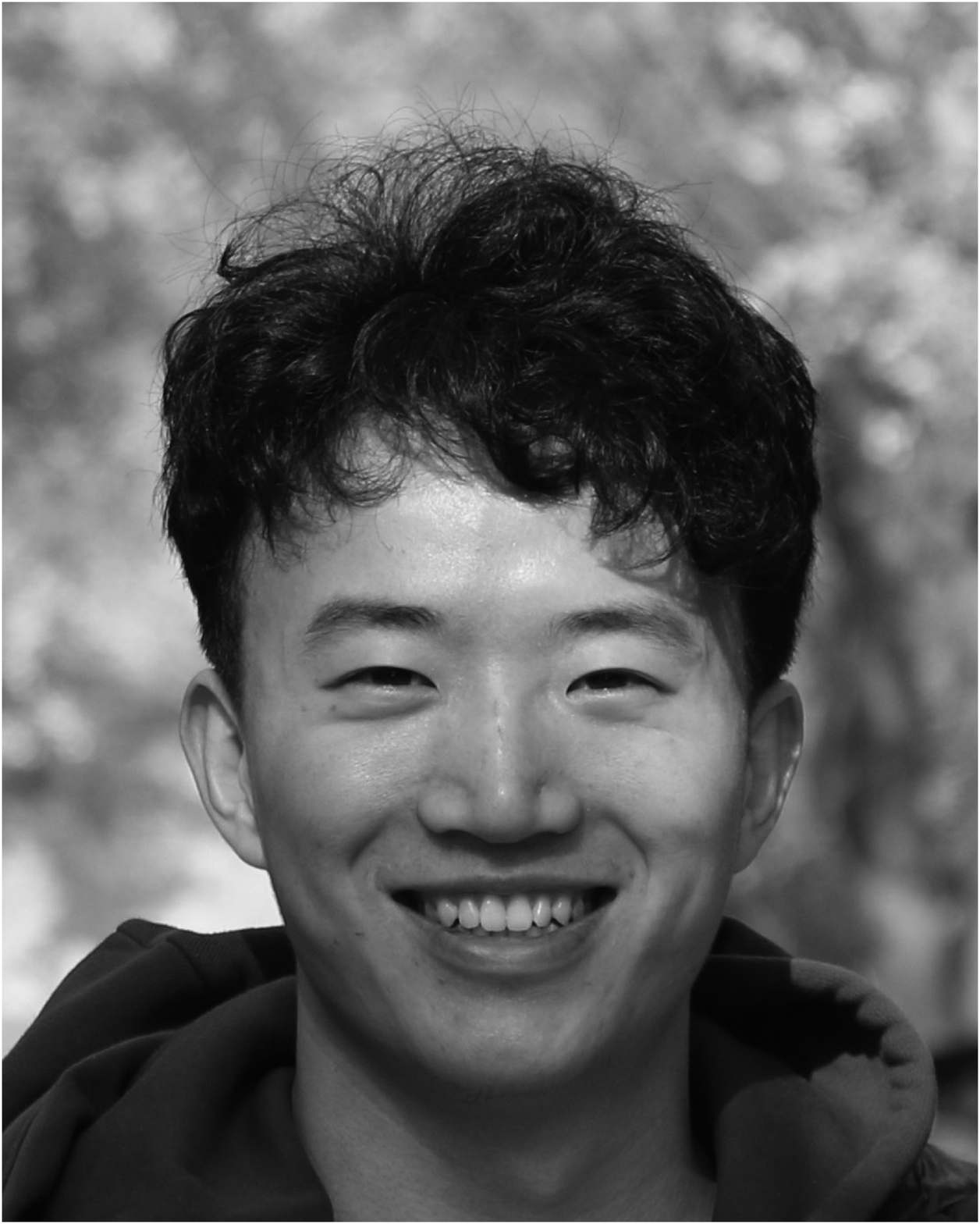}}]
{Kunjin Chen}received the B.Sc. degree in electrical engineering from Tsinghua University, Beijing, China, in 2015. Currently, he is a Ph.D. candidate with the Department of Electrical Engineering. 

His research interests include applications of machine learning and data science in power systems.
\end{IEEEbiography}

\vspace{-25pt}

\begin{IEEEbiography} [{\includegraphics[width=1in,height=1.25in,clip,keepaspectratio]{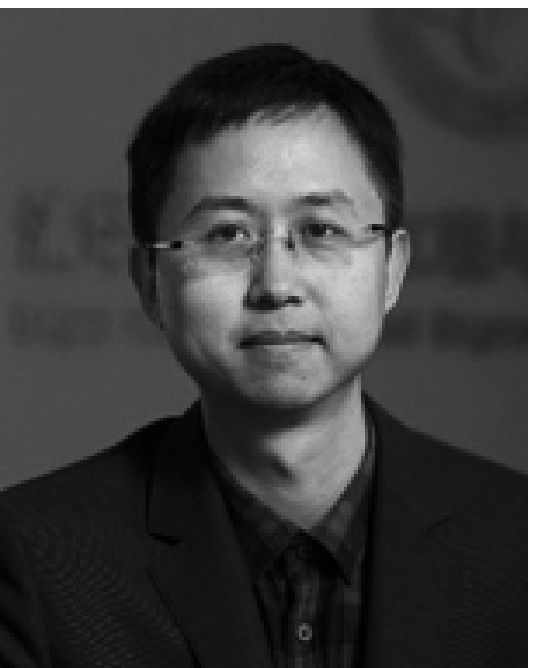}}]
{Jun Hu} (M'10) received his B.Sc., M.Sc., and Ph.D. degrees in electrical engineering from the Department of Electrical Engineering, Tsinghua University in Beijing, China, in July 1998, July 2000, July 2008. 

Currently, he is an associate professor in the same department. His research fields include overvoltage analysis in power system, sensors and big data, dielectric materials and surge arrester technology.
\end{IEEEbiography}

\vspace{-25pt}

\begin{IEEEbiography} [{\includegraphics[width=1in,height=1.25in,clip,keepaspectratio]{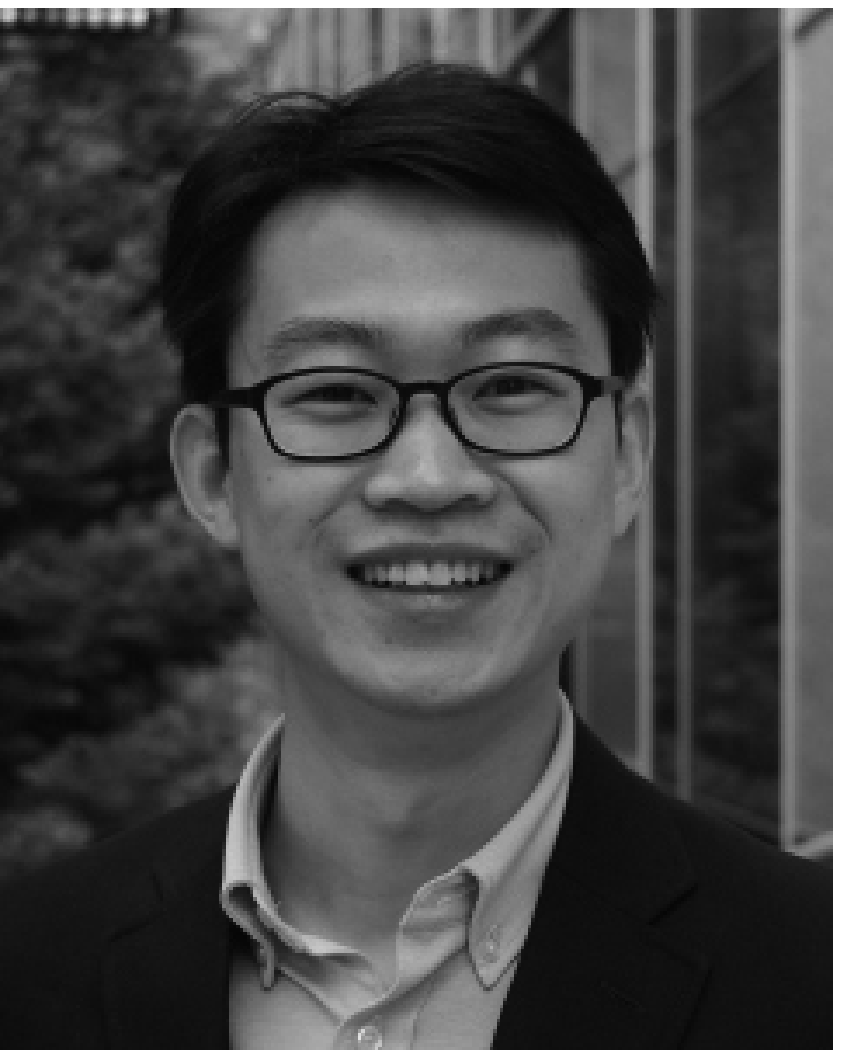}}]
{Yu Zhang} (M'15) received the Ph.D. degree in electrical and computer engineering from the University of Minnesota, Minneapolis, MN, USA, in 2015.

He is an Assistant Professor in the ECE Department of UC Santa Cruz. Prior to joining UCSC, he was a postdoc at UC Berkeley and Lawrence Berkeley National Laboratory. His research interests span the broad areas of cyber-physical systems, smart power grids, optimization theory, machine learning and big data analytics.
\end{IEEEbiography}

\vspace{-25pt}

\begin{IEEEbiography} [{\includegraphics[width=1in,height=1.25in,clip,keepaspectratio]{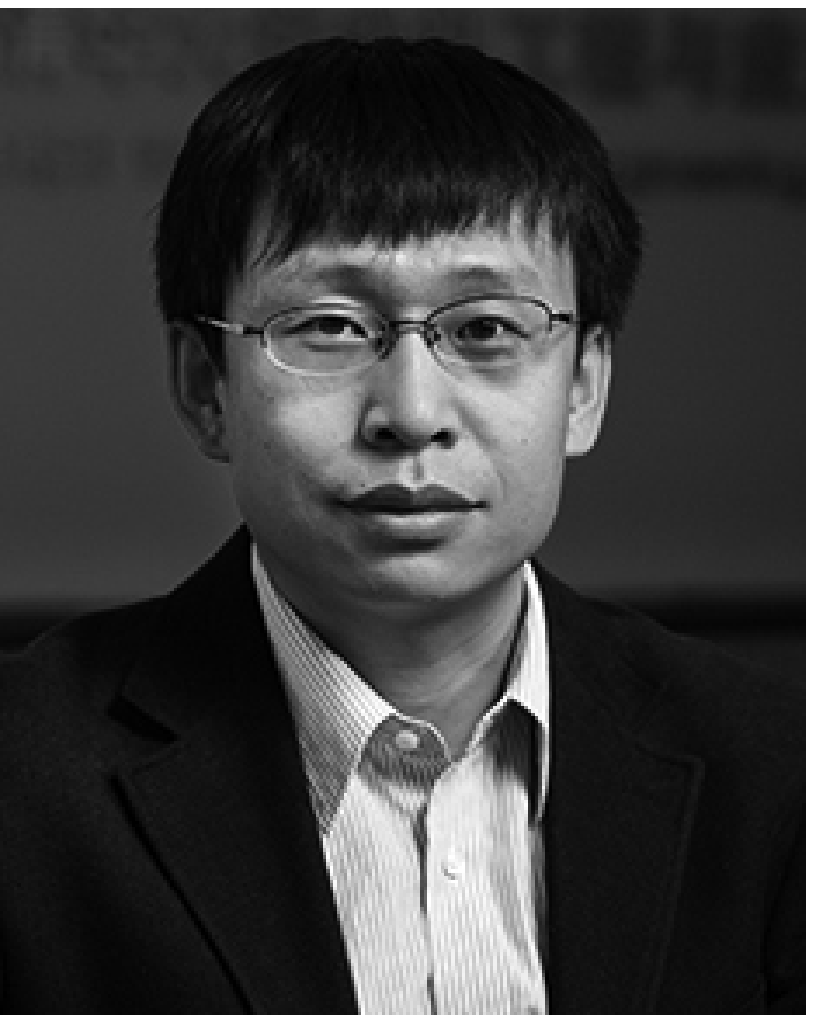}}]
{Zhanqing Yu} (M'07) received the B.Sc. and Ph.D. degrees in electrical engineering from Tsinghua University, Beijing, China, in 2003 and 2008, respectively.

He became a Postdoctoral Researcher in the Department of Electrical Engineering, Tsinghua University, Beijing, in 2008. He was a Lecturer in the same department from 2010 to 2012. Since 2013, he has been an Associate Professor. His research interests include electromagnetic compatibility and overvoltage analysis of HVDC and HVAC systems.
\end{IEEEbiography}

\vspace{-25pt}

\begin{IEEEbiography} [{\includegraphics[width=1in,height=1.25in,clip,keepaspectratio]{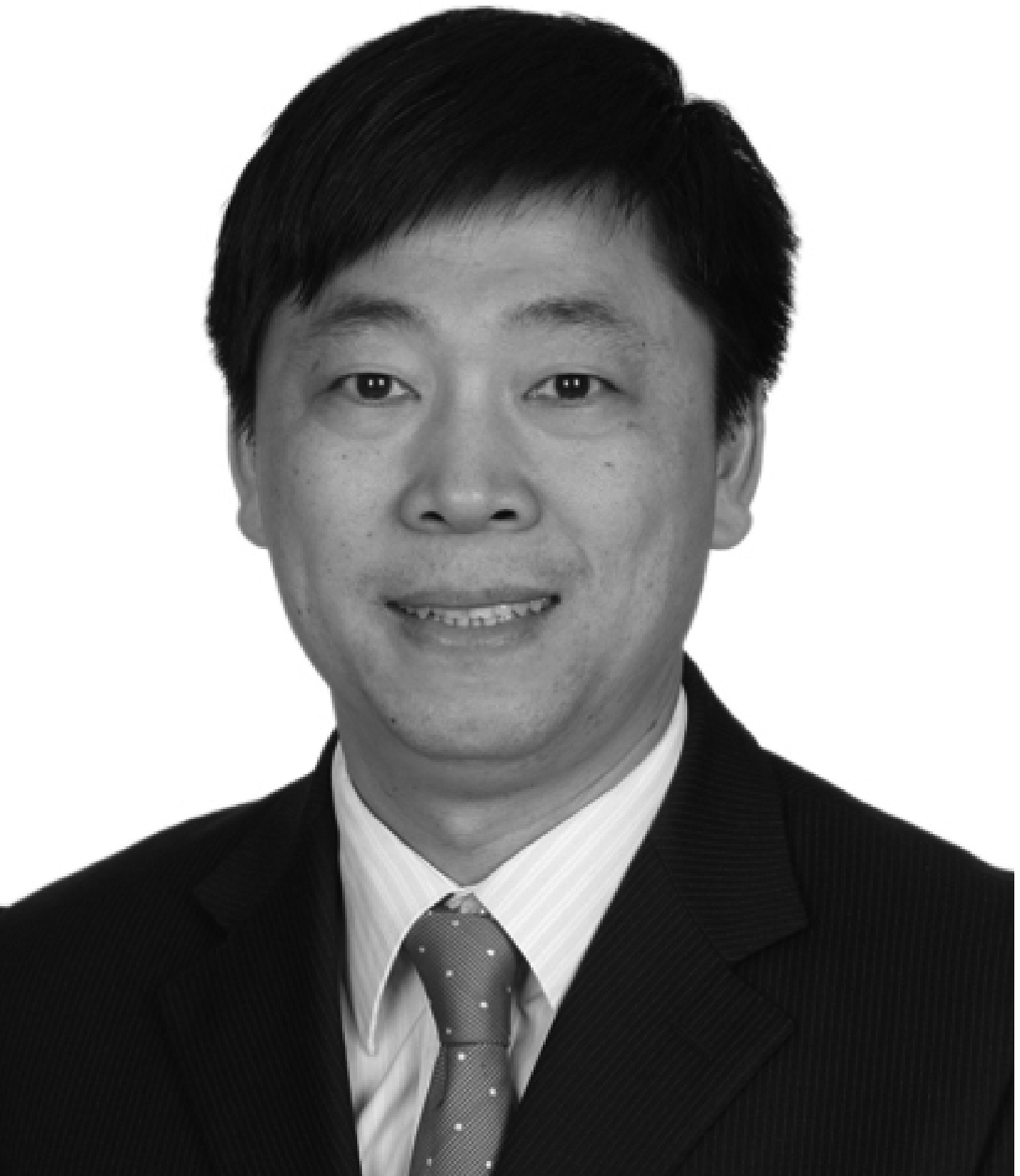}}]
{Jinliang He} (M'02--SM'02--F'08) received the B.Sc. degree from Wuhan University of Hydraulic and Electrical Engineering, Wuhan, China, the M.Sc. degree from Chongqing University, Chongqing, China, and the Ph.D. degree from Tsinghua University, Beijing, China, all in electrical engineering, in 1988, 1991 and 1994, respectively.

He became a Lecturer in 1994, and an Associate Professor in 1996, with the Department of Electrical Engineering, Tsinghua University. From 1997 to 1998, he was a Visiting Scientist with Korea Electrotechnology Research Institute, Changwon, South Korea, involved in research on metal oxide varistors and high voltage polymeric metal oxide surge arresters. From 2014 to 2015, he was a Visiting Professor with the Department of Electrical Engineering, Stanford University, Palo Alto, CA, USA. In 2001, he was promoted to a Professor with Tsinghua University. He is currently the Chair with High Voltage Research Institute, Tsinghua University. He has authored five books and 400 technical papers. His research interests include overvoltages and EMC in power systems and electronic systems, lightning protection, grounding technology, power apparatus, and dielectric material.
\end{IEEEbiography}
%
%\begin{IEEEbiography}[Kunjin-Chen.pdf]{Kunjin Chen}
%\end{IEEEbiography}

\end{document}